\PassOptionsToPackage{table}{xcolor}
\documentclass{adobe_research}
\usepackage[T1]{fontenc}
\usepackage{lmodern}
\usepackage{amsmath,amssymb}

\usepackage{amsmath,amsfonts,bm}

\def\eqref#1{equation~\ref{#1}}
\def\1{\bm{1}}

\DeclareMathAlphabet{\mathsfit}{\encodingdefault}{\sfdefault}{m}{sl}
\SetMathAlphabet{\mathsfit}{bold}{\encodingdefault}{\sfdefault}{bx}{n}

\usepackage{url}
\usepackage[table]{xcolor}
\usepackage{graphicx}
\usepackage{tikz}
\usepackage{tabularx,array}
\usepackage{adjustbox}
\usetikzlibrary{arrows.meta,positioning,fit,backgrounds,calc,shapes.geometric,shadows}
\usepackage{multirow}
\usepackage{subcaption}
\usepackage{microtype}
\usepackage{pifont}
\usepackage{wrapfig}
\usepackage{newtxtext}
\hypersetup{hypertexnames=false}

\graphicspath{{fig/}}

\definecolor{ccOursGreen}{HTML}{1F8A4F}   % Ours border + label
\definecolor{ccOursTint}{HTML}{EFF8F3}    % Ours row background
\definecolor{ccHeader}{HTML}{F3F5F8}      % prompt band background
\definecolor{ccCell}{HTML}{FFFFFF}        % prompt box fill
\definecolor{ccStripe}{HTML}{F7F8FA}      % spare, for a 3rd method row
\definecolor{ccRule}{HTML}{C9CCD1}        % rule under header
\definecolor{ccCellBd}{HTML}{D2D7DC}      % cell borders / row separators
\definecolor{ccOuterBd}{HTML}{A8B0BA}     % rounded outer frame
\definecolor{ccText}{HTML}{1F252D}        % prompt text
\definecolor{ccTextDim}{HTML}{6B7280}     % "Prompt" gutter label

\newcommand{\code}[1]{\texttt{\small #1}}

\newcommand{\designerrsititle}{Designer-RSI: Evolving Procedural Memory from User Traffic for Graphic Design Agent}
\title{\designerrsititle}

\renewcommand\author[2][]{\addtolist[#1]{#2}{\authorlist}{\authorformat}{\quad}}
\renewcommand\affiliation[2][]{\addtolist[#1]{#2}{\affiliationlist}{\affiliationformat}{\quad}}

\author[1,2]{Hongyang Du}
\author[1]{Lan Yan}
\author[1]{Christian Flores}
\author[1]{Asim Kadav}

\affiliation[1]{Adobe}
\affiliation[2]{Brown University}
\affiliation[]{Correspondence: \texttt{hongyang\_du@brown.edu}}

\abstract{\renewenvironment{abstract}{}{}\begin{abstract}
Deployed software agents face a continuous stream of user requests while their foundation-model weights remain frozen between releases. Recursive self-improvement (RSI) via external memory offers a practical path to continual adaptation, yet prior methods largely assume programmatic verifiers, cheap retries, or a single global prompt. Creative software breaks these assumptions, as open-ended visual briefs lack ground truth, VLM judges are noisy, and multi-step tool executions are costly and confounded by external assets. We present Designer-RSI, which evolves a modular \code{SKILL.md} skill bank from unlabeled design trajectories for an agent spanning over $230$ raster, vector, and layout tools. Designer-RSI couples \textbf{widening} (acquiring new skills for recurring capability gaps) with \textbf{deepening} (revising failing skills by contrasting successful and failed runs), and screens updates through a Pareto-style \textbf{matched replay gate} that requires strict gains on sampled briefs without local regressions. Across held-out product briefs, eight external benchmarks, and two backbones, traffic-driven evolution outperforms static documentation skills, lifts execution success up to $99.8\%$, improves requirement fulfillment, and reaches up to a $67.6\%$ blind win rate over the base agent. Ablations further show that skill acquisition and revision are strongly synergistic: neither suffices alone, and reliable gains require evolving coverage and execution quality together.
\end{abstract}}

\begin{document}
\definecolor{rc0}{HTML}{8A8F96}
\definecolor{rc1}{HTML}{7AD151}
\definecolor{rc2}{HTML}{22A884}
\definecolor{rc3}{HTML}{2A788E}
\definecolor{rc4}{HTML}{414487}
\definecolor{rc5}{HTML}{440154}
\maketitle
% Hide main-text sections from the ToC; only the supplementary is listed.
\addtocontents{toc}{\protect\setcounter{tocdepth}{-1}}

\section{Introduction}
\label{sec:intro}

Once deployed, software agents process a continuous stream of user requests that expose missing workflows, brittle tool sequences, and recurring failure modes. Because frontier foundation models remain frozen between releases, deployed agents repeat the same mistakes across sessions. Recursive self-improvement (RSI)~\citep{good1965speculations,zelikman2024stop,zhang2025darwingodelmachine} offers a direct way to close this loop: an agent executes tasks, distills operational lessons from its trajectories, and uses the updated capabilities to serve subsequent traffic. Despite growing interest, practical demonstrations of RSI remain largely confined to sandboxed environments with clean programmatic verifiers, such as code generation with unit tests~\citep{zelikman2024stop,zhang2025darwingodelmachine}, game worlds with explicit state checks~\citep{wang2023voyager}, or short-horizon tasks where same-task retries are cheap~\citep{shinn2023reflexion,zhao2024expel,wang2025awm}. In creative software, live traffic lacks ground-truth outputs and human labels, while unverified updates risk breaking live workflows.

\begin{figure*}[t]
\centering
\includegraphics[width=\linewidth]{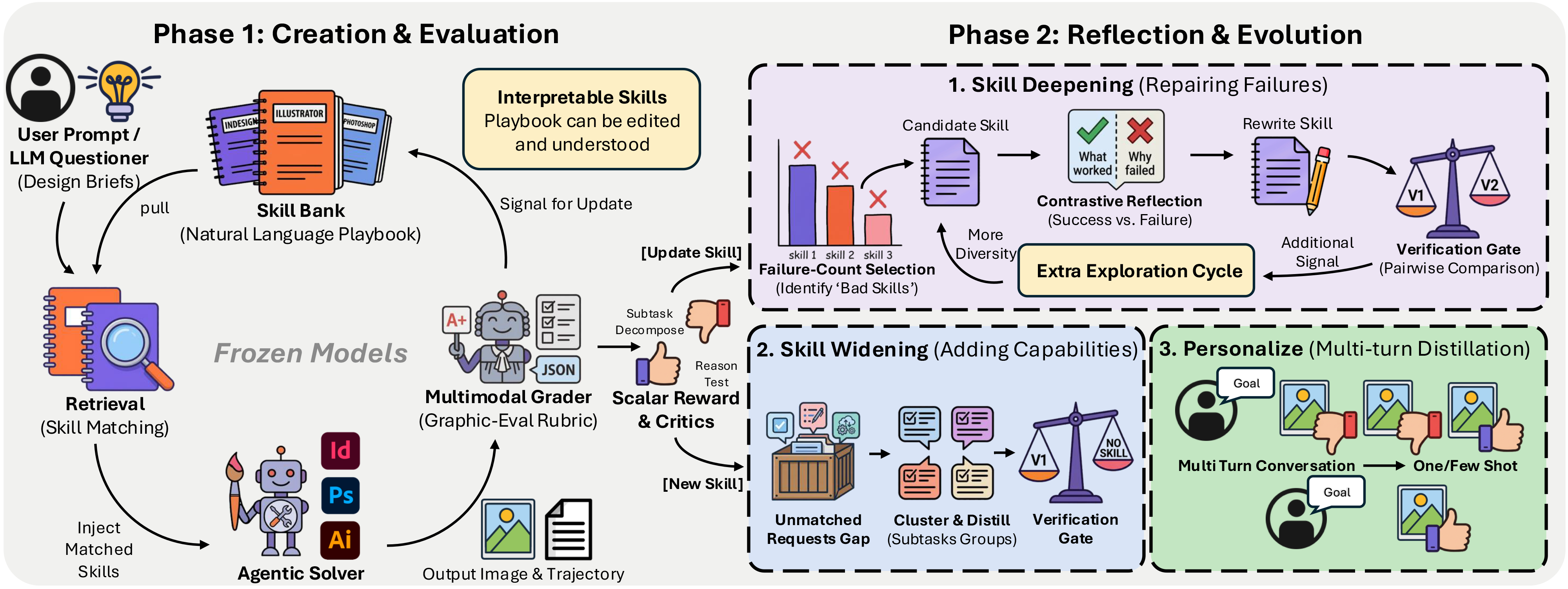}
\caption{\textbf{Traffic-driven RSI through procedural memory.} \emph{Phase 1:} The frozen design agent retrieves relevant skills from a versioned \code{SKILL.md} bank to execute user-derived and LLM-augmented briefs, while an offline multimodal grader (\code{Graphic-Eval}) produces requirement-level scores and failure rationales. \emph{Phase 2:} \textbf{Deepening} revises failing skills via contrastive success--failure reflection, \textbf{widening} distills new skills from recurring uncovered subtasks, and a \textbf{matched replay gate} screens candidate updates pairwise before committing them to the bank. Panel 3 illustrates a separate user-guided personalization extension (Appendix~\ref{app:personalization}).}
\label{fig:teaser}
\end{figure*}

Professional graphic design shows why complex creative systems resist both static prompt engineering and off-the-shelf memory evolution. Turning an open-ended brief into an editable, multi-layer document~\citep{jia2023cole,Inoue_2024_CVPR,hong2026creatipostereditablecontrollablemultilayer} requires orchestrating over $230$ tools across photo editing, vector illustration, and page layout~\citep{wang-etal-2025-banneragency,ki2025graphicbenchplanningbenchmarkgraphic}. Upfront manual curation falls short: workflows distilled from product documentation fail to outperform a no-skill baseline because help articles explain individual features in isolation rather than end-to-end workflows. At the same time, design trajectories break standard assumptions in agent memory learning~\citep{zhao2024expel,wang2025awm,ouyang2025reasoningbank,zhang2026agentic,zhou2026mementoskills,yang2026skillopt}: visual outputs lack programmatic oracles while VLM judges exhibit run-to-run drift and order bias~\citep{creval,chang2025oneig,zheng2023judging,wang2024fair} confounded by retrieved stock assets; capabilities span hundreds of tools that must be retrieved selectively rather than stuffed into one prompt; and single sessions run up to $15$ minutes and hundreds of thousands of tokens, ruling out online trial-and-error retries.

To make RSI work in this setting, we build Designer-RSI around an external, human-auditable \textbf{skill bank} of modular \code{SKILL.md} workflows while keeping all foundation models frozen (Figure~\ref{fig:teaser}). Our offline evolution loop rests on two principles. First, procedural memory evolves along two coupled axes: \textbf{widening} expands coverage by distilling new skills when uncovered subtasks recur across requests, while \textbf{deepening} improves reliability by revising failure-prone skills through contrasts between their successful and failed executions. Second, we decouple open-ended proposal from conservative admission: every candidate addition or revision must pass a \textbf{matched replay gate}~\citep{thomas2015highconfidence,laroche2019safe} that compares candidate and baseline arms under identical retrieved asset pools and document state, committing an update only when it improves at least one sampled brief and regresses on none.

Across five rounds of evolution, Designer-RSI grows a documentation-seeded bank into multi-tool workflows that substantially improve both task completion and visual quality across model scales. Controlled ablations further reveal why expansion and refinement cannot be decoupled: newly acquired skills start as rough first-draft procedures induced from a handful of traces, and require iterative trajectory-grounded revision to execute reliably alongside the rest of the bank. Our contributions are:
\begin{itemize}
  \item \textbf{RSI for open-ended creative tool use.} We demonstrate recursive self-improvement in an industrial-scale graphic-design agent, turning unlabeled execution trajectories into modular procedural memory without ground-truth verifiers, human annotations, or weight updates.
  
  \item \textbf{Coupled skill evolution with matched replay.} We pair recurrence-gated skill acquisition with contrastive success--failure revision, screened by a matched A/B replay gate that controls for external asset variance and blocks locally regressive updates under noisy VLM judging.
  
  \item \textbf{Multi-benchmark evaluation and ablation analysis.} Across held-out product briefs, eight external design and compositional benchmarks, and two model backbones, we show consistent gains in execution completion, requirement fulfillment, and blind visual preference, diagnose why static documentation skills fall short, and establish the synergy between skill acquisition and revision.
\end{itemize}

\section{Related Work}
\label{sec:related}

\begin{table*}[t]
\centering
\caption{\textbf{Comparison with prior external-memory and skill-evolution frameworks (frozen LLM weights).} Prior methods update a single prompt, playbook, or skill, rely on ground-truth (GT) checks or benchmark scores, or accept edits via direct writes, counters, single-run filters, or one synthetic unit test. Designer-RSI adds and revises global \code{SKILL.md} workflows (plus user-guided personal skills, Appendix~\ref{app:personalization}) using a reference-free VLM judge and matched A/B replay.}
\label{tab:related_work}
\begingroup
\small
\setlength{\tabcolsep}{5pt}
\renewcommand{\arraystretch}{1.12}
\begin{tabularx}{\textwidth}{@{}ll>{\raggedright\arraybackslash}Xll@{}}
\toprule
\textbf{Method} & \textbf{Memory structure} & \textbf{Update ops} & \textbf{Feedback signal} & \textbf{Update check} \\
\midrule
Voyager~\citep{wang2023voyager}
& Code skill bank & Add skills & Env.\ check (GT) & Single-run pass \\
ExpeL~\citep{zhao2024expel}
& Single rule list & Add, revise, prune rules & Task pass/fail (GT) & Vote counters \\
AWM~\citep{wang2025awm}
& Text workflow list & Add workflows & Text LLM judge (no GT) & Single-run filter \\
ReasoningBank~\citep{ouyang2025reasoningbank}
& Text memory bank & Add, merge items & Text LLM judge (no GT) & None (direct write) \\
ACE~\citep{zhang2026agentic}
& Single text playbook & Add, revise, prune rules & GT labels & Counters \& dedup \\
GEPA~\citep{agrawal2025gepa}
& Single system prompt & Revise, merge prompt & Benchmark score (GT) & Val-set score check \\
SkillOpt~\citep{yang2026skillopt}
& Single \code{SKILL.md} & Revise single skill & Benchmark score (GT) & Val-set score check \\
Memento-Skills~\citep{zhou2026mementoskills}
& \code{SKILL.md} bank & Add, revise global skills & Answer match (GT) & 1 synthetic unit test \\
\midrule
\rowcolor{black!5}
\textbf{Designer-RSI} (Ours)
& \textbf{\code{SKILL.md} bank}
& \textbf{Add, revise global \& personal skills}
& \textbf{VLM rubric judge (no GT)}
& \textbf{Matched A/B replay} \\
\bottomrule
\end{tabularx}
\endgroup
\end{table*}

\paragraph{Self-improving agents and experiential memory.}
Recent language agents instantiate recursive self-improvement by modifying external artifacts rather than model weights, mainly under programmatic verifiers or cheap retries (Table~\ref{tab:related_work}): STOP~\citep{zelikman2024stop} and the Darwin G\"odel Machine~\citep{zhang2025darwingodelmachine} rewrite code against unit tests, Voyager~\citep{wang2023voyager} saves Minecraft skills after one state check, and Reflexion~\citep{shinn2023reflexion} carries critiques across same-task retries. Closer to our setting, ExpeL~\citep{zhao2024expel} curates insights from binary rewards via vote counters, AWM~\citep{wang2025awm} induces workflows from judged successes, and ReasoningBank~\citep{ouyang2025reasoningbank} merges reasoning items across outcomes, yet none revises procedures in-place via A/B replay.

\paragraph{Prompt, context, and skill optimization.}
A parallel line optimizes instructions with frozen weights: TextGrad~\citep{yuksekgonul2024textgrad} backpropagates critiques, Dynamic Cheatsheet~\citep{suzgun2026dynamiccheatsheet} curates test-time notes, GEPA~\citep{agrawal2025gepa} evolves one prompt via a Pareto archive, and ACE~\citep{zhang2026agentic} curates a playbook via counters and deduplication. Among prior \code{SKILL.md} methods, SkillOpt~\citep{yang2026skillopt} tunes one skill file against held-out benchmark scores, while Memento-Skills~\citep{zhou2026mementoskills} evolves \code{SKILL.md}-and-script folders with an RL-trained router using ground-truth answer checks, online retries, single-trace patching, and one synthetic unit test. Concurrent work also studies skill evolution: SkillAudit~\citep{gao2026skillaudit} revises an existing skill via paired trajectory audits without acquiring new skills or matching external assets, and EvoSkill~v2~\citep{huang2026evoskillv2} rewrites coding skills via a stateful agent against a benchmark grader without non-regression gating. By contrast, Designer-RSI keeps skill retrieval training-free while widening and deepening a cross-application \code{SKILL.md} bank on unlabeled trajectories via matched replay.

\paragraph{Why production graphic design requires a different loop.}
Open-ended design traffic breaks the assumptions along every column of Table~\ref{tab:related_work}:
(i)~\emph{Memory scale}: spanning hundreds of cross-application tools requires a selectively retrieved multi-skill bank beyond what one global prompt, list, or skill file can hold;
(ii)~\emph{Coupled evolution}: traffic exposes both missing workflows and brittle seed skills, requiring recurrence-gated acquisition (\emph{widening}) and contrastive in-place revision (\emph{deepening});
(iii)~\emph{Reference-free visual grading}: briefs lack ground truth and scalar VLM scores drift across runs (Appendix~\ref{app:gate}), ruling out ground-truth checks and raw score thresholds; and
(iv)~\emph{Confounded, costly rollouts}: retrieved stock assets shift task difficulty and multi-step sessions preclude online retries, requiring localized head-to-head replay under matched context over unscreened updates, synthetic unit tests, or benchmark sweeps.

\paragraph{Editable graphic design and VLM evaluation.}
COLE~\citep{jia2023cole}, OpenCOLE~\citep{Inoue_2024_CVPR}, CreatiPoster~\citep{hong2026creatipostereditablecontrollablemultilayer}, and CreatiDesign~\citep{zhang2026creatidesign} generate layered graphic layouts, while BannerAgency~\citep{wang-etal-2025-banneragency} and GraphicBench~\citep{ki2025graphicbenchplanningbenchmarkgraphic} study multimodal design planning and tool use. Because VLM judges exhibit order bias~\citep{wang2024fair,zheng2023judging}, we use position-swapped pairwise comparisons. Our replay gate adapts baseline bootstrapping from safe policy improvement~\citep{laroche2019safe,thomas2015highconfidence} as a practical non-regression screen.

% Figure 2: Image 2 pipeline using the overview's blue/lavender/yellow/gray palette.
% The original editable implementation is retained in fig_pipelines_tikz.tex.
\begin{figure*}[t]
\centering
\includegraphics[width=\linewidth]{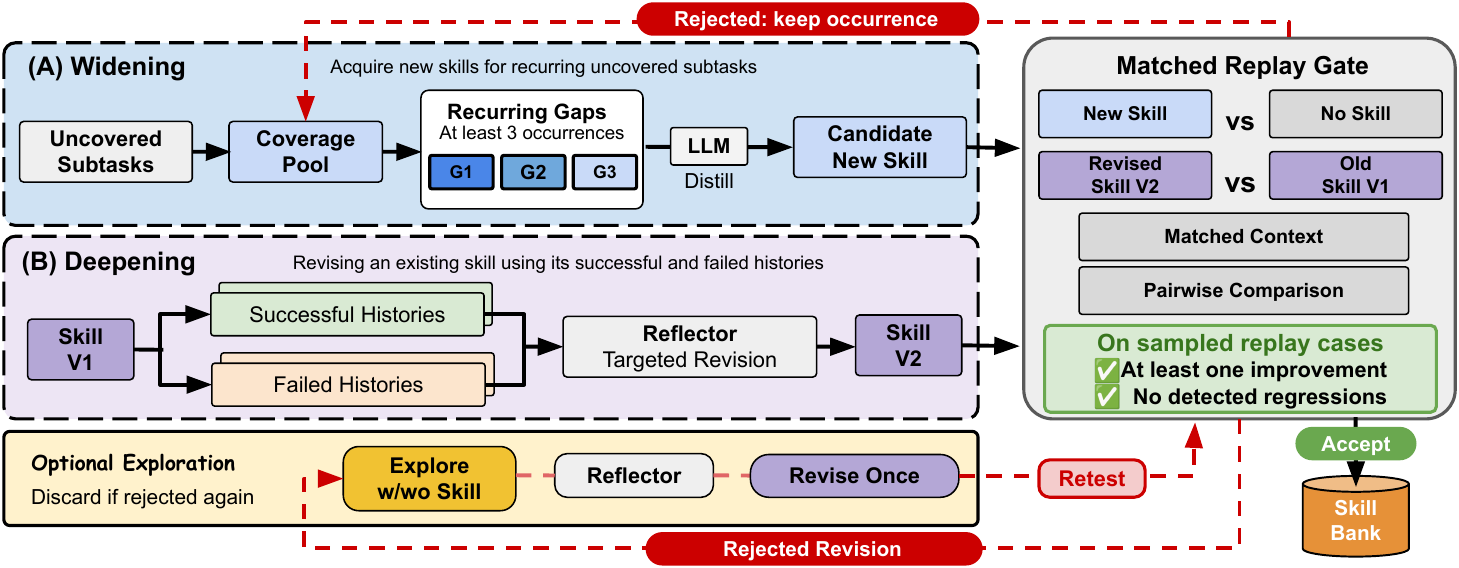}
\caption{\textbf{Skill-bank update mechanisms and admission screening.}
\textbf{(A) Widening} clusters uncovered subtasks into recurring capability gaps, distills a candidate \code{SKILL.md}, and tests it against the no-skill baseline. Rejected candidates leave their occurrences in the pool for later rounds.
\textbf{(B) Deepening} contrasts a failing skill's successful and failed trajectories to propose a targeted edit or full rewrite, tested against the incumbent on stratified histories. A rejected rewrite triggers one diagnostic exploration cycle.}
\label{fig:cluster}\label{fig:rewrite}
\end{figure*}

\section{Graphic Design Agent and Skill Bank}
\label{sec:agent-background}

Our agent constructs structured, editable design documents through more than $230$ tools spanning raster editing, vector graphics, and multi-page layout, covering workflows comparable to Adobe Photoshop, Illustrator, and InDesign (Appendix~\ref{app:agent_detail}). Controlled by a frozen frontier LLM, the agent retrieves licensed stock assets, executes multi-step tool calls, and inspects intermediate renders before finalizing a design. Every document state serializes to deterministic XML, so completed trajectories can be re-rendered and graded offline without slowing live requests.

\paragraph{Skill bank and progressive retrieval.}
Procedural memory is stored as a versioned bank of Markdown files (\code{SKILL.md}), partitioned by application mode (Appendix~\ref{sec:bank}). Each skill's YAML front matter defines an operation-centric trigger \code{description} and critical \code{tool\_sequence}, while its body specifies applicability rules, step-by-step tool parameters, domain tips, and error recovery. Because user briefs are verbose and subject-heavy whereas skills are operation-centric, retrieval runs in two stages: a frozen LLM first distills the brief into at most six core operations (stripping subject nouns and style adjectives), and lexical token-overlap scoring flags the top $2$--$4$ matching skills in a compact catalog appended to the system prompt. To keep the context compact, the agent selectively invokes \code{load\_skill(name)} and reads only the procedures it chooses to apply (optional tool-list narrowing is disabled in our main experiments). Disabling skill access yields the no-skill baseline (\textsc{Base}).

\paragraph{Trajectory logging and cold-start seeding.}
For every execution, the runtime logs the brief, distilled query, retrieved skills, tool-call trace, and final XML snapshot. Offline \code{Graphic-Eval} grading appends requirement-level outcomes and failure rationales (Appendix~\ref{sec:eval}). Before traffic-driven evolution begins, a four-phase pipeline seeds the bank from official product help articles, producing $76$ initial skills (\textsc{Cold Start}; Appendix~\ref{sec:coldstart}).

% \input{sections/skillbank}

% Ensure at most one top float per page from Section 4 through Conclusion.
\setcounter{topnumber}{1}
\setcounter{dbltopnumber}{1}

\section{Evolving Procedural Memory from Traffic}
\label{sec:evolve}

\begin{wraptable}{r}{0pt}
\small
\begin{tabular}{@{}ll@{}}
\toprule
\textbf{Component} & \textbf{Function in the evolution loop} \\
\midrule
\textbf{Prompter} & user-derived briefs and LLM-augmented variants \\
\textbf{Solver} & executes briefs using the skill bank and toolset \\
\textbf{Grader} & scores outputs and explains unmet requirements \\
\textbf{Reflector} & distills new skills and revises failing \code{SKILL.md} \\
\bottomrule
\end{tabular}
\end{wraptable}
Our offline loop turns recorded design trajectories into persistent updates to the skill bank (Figure~\ref{fig:teaser}). In each round, the current agent executes a batch of briefs derived from historical user traffic and LLM-augmented variants, an automated grader scores the rendered outputs, and a reflector proposes skill additions or revisions that are screened before entering the next round's bank. Throughout evolution, all foundation models, tools, rendering, grading, and update rules remain frozen. Only the \code{SKILL.md} files change.

Two complementary operations propose updates (Figure~\ref{fig:cluster}): \textbf{widening} acquires skills for recurring capability gaps, and \textbf{deepening} revises existing skills with repeated failures. Both pass through a shared matched replay gate. User-guided personalization is studied in Appendix~\ref{app:personalization} and excluded from the shared bank and main experiments.

\begin{figure*}[t]
  \centering
  \begin{subfigure}[t]{0.49\linewidth}
    \centering
    \includegraphics[width=\linewidth]{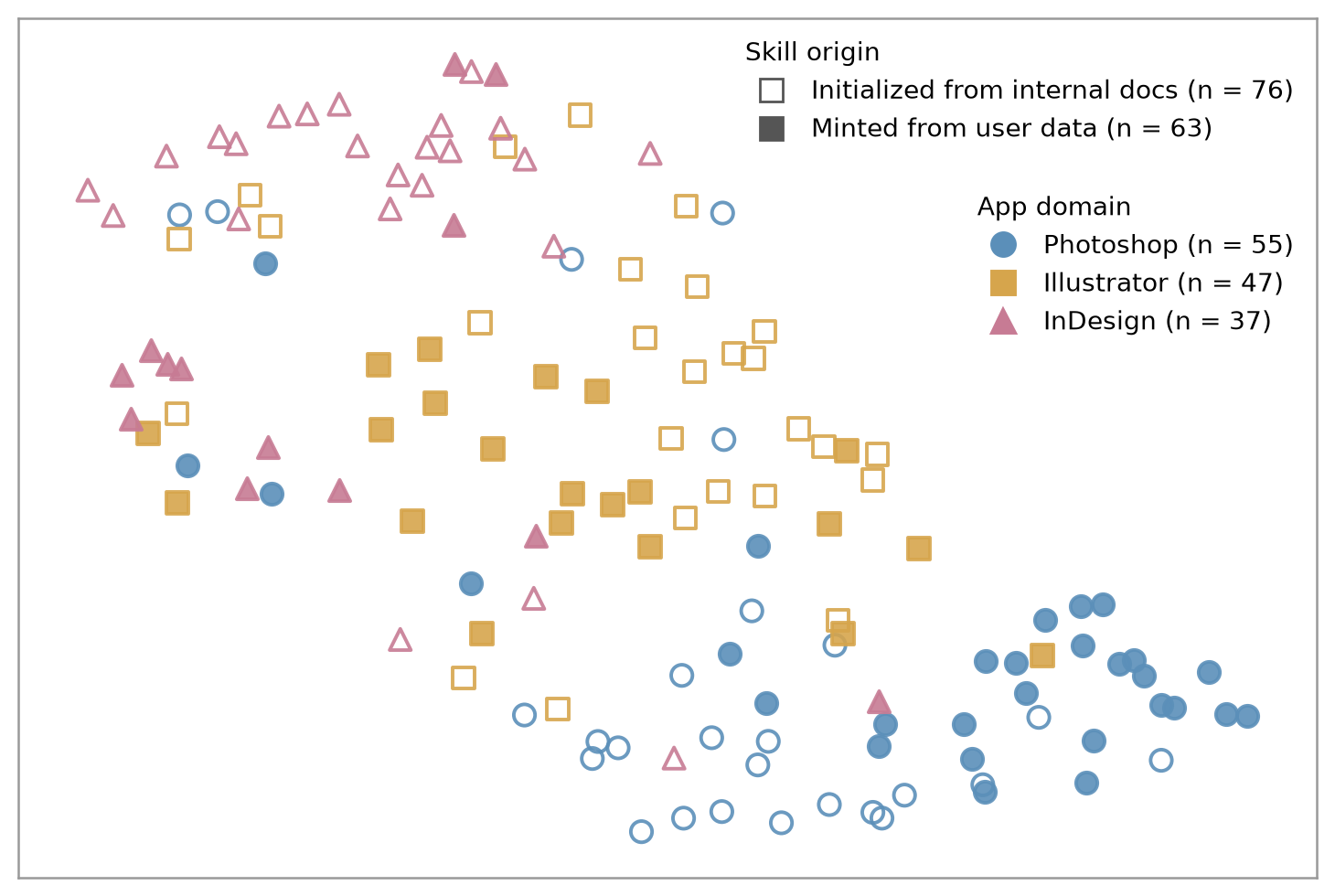}
    \caption{Skill embedding}
    \label{fig:skill-umap}
  \end{subfigure}\hfill
  \begin{subfigure}[t]{0.49\linewidth}
    \centering
    \includegraphics[width=\linewidth]{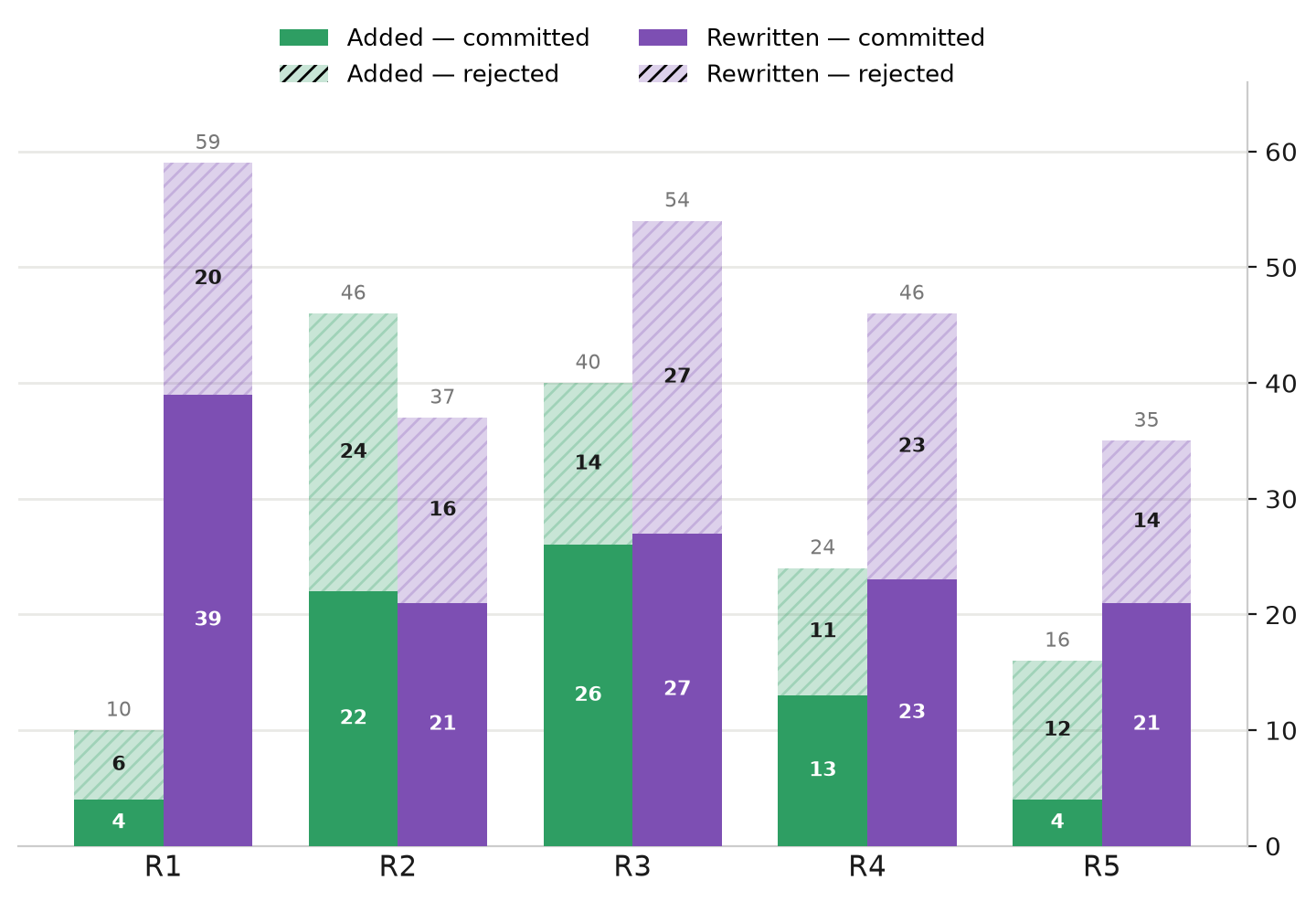}
    \caption{Skill changes}
    \label{fig:skill-flow}
  \end{subfigure}
  \caption{\textbf{Skill-bank evolution across five rounds.}
  (a)~UMAP of skill descriptions (\texttt{BGE-small-en-v1.5}, cosine) across Photoshop (PS), Illustrator (AI), and InDesign (ID) for $76$ seed (hollow) and $63$ newly acquired (filled) skills.
  (b)~Per-round skill additions and revisions: accepted (solid) vs.\ rejected (hatched).}
  \label{fig:skills}
\end{figure*}

\subsection{Widening: Acquiring New Skills}
\label{sec:gapcluster}

After each execution, a frozen LLM extracts and canonicalizes the subtasks performed from the brief and tool-call trace. A subtask is marked \emph{uncovered} when retrieval returns no match, when retrieved skills cover only other parts of the brief, or when an existing skill repeatedly fails to resolve it during deepening (\S\ref{sec:deepening}). Uncovered subtasks accumulate in a persistent coverage pool indexed by canonical label. Once a label reaches $k_{\min}=3$ occurrences, the Reflector distills those trajectories into a candidate \code{SKILL.md}. Requiring recurrence prevents one-off prompts from bloating the bank and focuses acquisition on recurring demand. The candidate commits only if it passes the matched replay gate against the no-skill \textsc{Base} agent (\S\ref{sec:gate}). If rejected, the candidate is discarded while its underlying occurrences stay in the pool so future rounds can revisit with richer evidence.

\subsection{Deepening: Revising Existing Skills}
\label{sec:deepening}

Whereas widening expands coverage, deepening repairs skills that execute unreliably. Each graded trajectory receives a composite score $s_j \in [0,1]$ weighting requirement completeness ($0.7$) and visual aesthetics ($0.3$) (Appendix~\ref{app:scoring}). Any trajectory below the failure cutoff ($s_j < 0.6$) increments the failure counter of every skill it retrieved. Absolute VLM scores only prioritize candidates here: skills reaching at least $m=2$ failures are queued for revision (most-failing first), while actual admission relies exclusively on pairwise replay.

\begin{figure*}[t]
\centering
%======== 左:图(更宽)========
\begin{subfigure}[c]{0.49\linewidth}
  \centering
  \includegraphics[width=\linewidth]{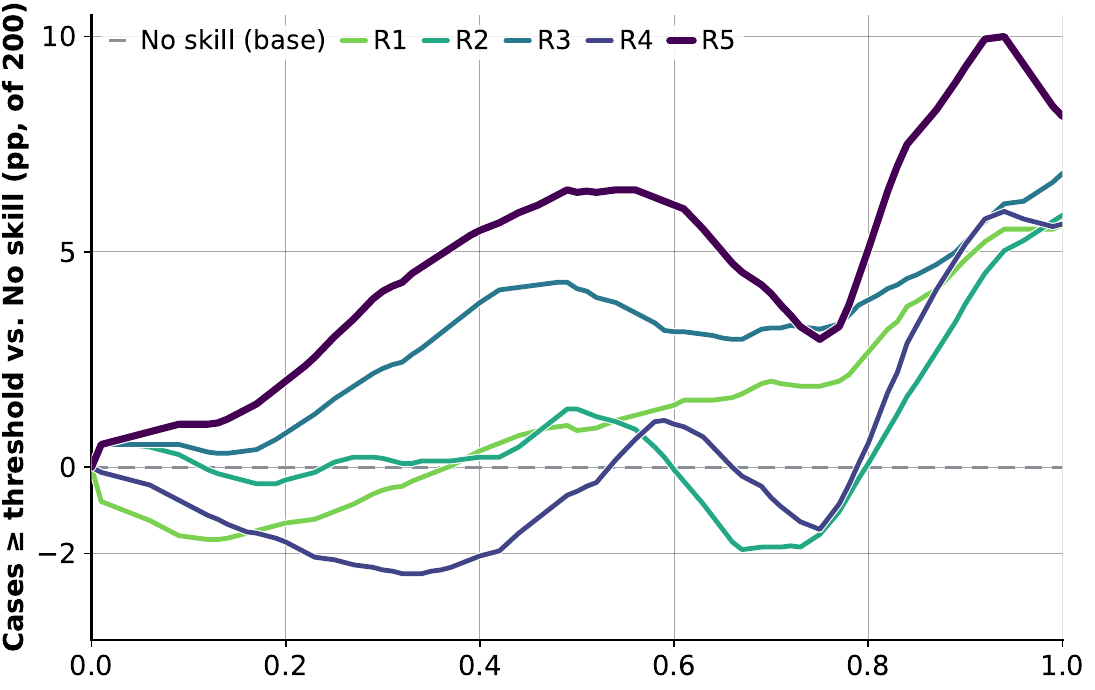}
  \caption{Survival gap}
  \label{fig:retention-diff}
\end{subfigure}
\hfill
%======== 右:表 ========
\begin{subfigure}[c]{0.49\linewidth}
  \centering
  \small\setlength{\tabcolsep}{3pt}
  \begin{tabular}{lcccccc}
  \toprule
  $\ge\tau$ & \cellcolor{rc0}\textcolor{white}{\textbf{Base}} & \cellcolor{rc1}\textcolor{white}{\textbf{R1}} & \cellcolor{rc2}\textcolor{white}{\textbf{R2}} & \cellcolor{rc3}\textcolor{white}{\textbf{R3}} & \cellcolor{rc4}\textcolor{white}{\textbf{R4}} & \cellcolor{rc5}\textcolor{white}{\textbf{R5}} \\
  \midrule
  $\ge 0.1$ & 96 & 95 & \textbf{98} & 97 & 96 & \textbf{98} \\
  $\ge 0.2$ & 96 & 94 & 96 & 96 & 94 & \textbf{98} \\
  $\ge 0.3$ & 94 & 92 & 94 & 96 & 90 & \textbf{98} \\
  $\ge 0.4$ & 91 & 91 & 90 & 94 & 88 & \textbf{96} \\
  $\ge 0.5$ & 86 & 88 & 88 & 92 & 85 & \textbf{93} \\
  $\ge 0.6$ & 80 & 82 & 80 & 83 & 80 & \textbf{86} \\
  $\ge 0.7$ & 74 & 76 & 71 & 76 & 73 & \textbf{78} \\
  $\ge 0.8$ & 62 & 64 & 62 & \textbf{66} & 62 & 64 \\
  $\ge 0.9$ & 43 & 48 & 46 & 48 & 50 & \textbf{56} \\
  $=1.0$    & 24 & 30 & 30 & 31 & 30 & \textbf{32} \\
  \bottomrule
  \end{tabular}
  \caption{Retention (\%)}
  \label{fig:retention}
\end{subfigure}
  \caption{\textbf{Held-out completeness across five rounds} ($200$ briefs, \texttt{Claude-Sonnet-4}).
  (a)~Completeness survival gain $\Delta R(\tau)$ over \textsc{Base} across thresholds $\tau \in [0,1]$.
  (b)~Retention $R(\tau)$: percentage of briefs reaching completeness $\ge \tau$ (row maxima in \textbf{bold}).}
  \label{fig:evolution-overview}
\end{figure*}

To revise a selected skill, the Reflector inspects its current \code{SKILL.md} alongside contrasting execution histories: failing briefs paired with unmet-requirement rationales versus successful trajectories of the \emph{same} skill (including tool sequences, intermediate render checks, and agent reasoning). Comparing the two outcomes anchors each edit in concrete tool calls and parameters, which repairs failure modes while preserving working instructions. The Reflector first proposes a targeted edit. If repeated targeted edits fail the replay gate, it escalates to a full rewrite.

If a full rewrite also fails, the loop runs one diagnostic exploration cycle, probing failing briefs with and without the skill to guide one final revision and decide whether to update, deprecate, or retain it. Unresolved failures from a retained skill return to the widening coverage pool. Widening and deepening thus reinforce each other: newly acquired skills are refined as they accumulate usage, while persistent failures surface missing sub-procedures for acquisition.

\subsection{Matched Replay Gate}
\label{sec:gate}

Admitting updates directly from scalar VLM scores is brittle: a VLM Grader's absolute score drifts across runs, and upstream stochasticity (especially which stock assets are retrieved) can make one run much easier than another. Our replay gate removes both confounders by comparing candidate and baseline arms head-to-head under \emph{matched upstream context}: deepening tests the revision against the incumbent skill, while widening tests the new skill against \textsc{Base}.

For each candidate update, we sample briefs that exercise the skill (from the triggering cluster for widening, or stratified across past successes and failures for deepening) and prepare multiple upstream contexts (fixed retrieved asset pools and initial document state) per brief. Within each context, candidate and baseline arms execute in the same batch from identical inputs, and a VLM judge compares their rendered outputs across swapped presentation orders with a tie band to absorb minor judge jitter (Appendix~\ref{app:gate}). A sampled brief counts as won (or lost) when the candidate wins (or loses) a majority of its contexts beyond the tie band. An update commits if and only if it satisfies a Pareto-style dominance condition across sampled briefs:
\begin{equation}
  \bigl(\nexists\text{ sampled brief lost}\bigr)\ \wedge\
  \bigl(\exists\text{ sampled brief won}\bigr).
  \label{eq:accept-main}
\end{equation}
During deepening, sampling both past successes and past failures checks that a revision fixes broken runs without regressing on cases the incumbent already handled. Because Eq.~(\ref{eq:accept-main}) screens updates locally over sampled briefs, we evaluate whole-bank progress separately on held-out briefs.

\section{Experiments}
\label{sec:experiments}
\begin{table*}[t]
\centering\small
\caption{\textbf{Generalization across eight external benchmarks on two frozen backbones} ($300$ prompts per benchmark).
\textsc{Evolve} uses the round-5 skill bank evolved by Designer-RSI, while \textsc{Base} disables skill access.
\textbf{(a)}~Execution success rates (\textsc{SR}, with $\text{\textsc{Evolve}}-\text{\textsc{Base}}$ gain in parentheses) and blind pairwise win rates over \textsc{Base} (\texttt{GPT-5.4}, position-swapped).
\textbf{(b)}~General prompt-following transfer: quality scores on completed runs with \textsc{SR} (\%) in parentheses ($\Delta = \text{\textsc{Evolve}} - \text{\textsc{Base}}$).}
\label{tab:main-results}
\setlength{\tabcolsep}{4pt}
\renewcommand{\arraystretch}{1.13}
\begin{tabular*}{\linewidth}{@{\extracolsep{\fill}}ll cccc >{\columncolor{black!3}[\tabcolsep][0pt]}c@{}}
\toprule
\rowcolor{black!6}
\multicolumn{7}{l}{\textbf{(a) Specialized graphic-design benchmarks:  execution success (\%) and pairwise win rate vs.\ \textsc{Base} }} \\
\textbf{Backbone} & \textbf{Metric} & \textbf{OpenCOLE}\,$\uparrow$ & \textbf{GraphicBench}\,$\uparrow$
 & \textbf{CreatiDesign}\,$\uparrow$ & \textbf{\scriptsize BannerRequest400}\,$\uparrow$ & \textbf{Avg}\,$\uparrow$ \\
\midrule
\multirow{3}{*}{\text{Claude-Opus-4.6}}

  & \textsc{Base} SR  & \textbf{100.0\%} & 98.7\% & 94.7\% & \textbf{100.0\%} & 98.3\% \\
  & \textsc{Evolve} SR & \textbf{100.0\%}~{\textcolor{gray}{\scriptsize (0.0)}} & \textbf{100.0\%}~{\textcolor{green!60!black}{\scriptsize (+1.3)}} & \textbf{96.7\%}~{\textcolor{green!60!black}{\scriptsize (+2.0)}} & \textbf{100.0\%}~{\textcolor{gray}{\scriptsize (0.0)}} & \textbf{99.2\%}~{\textcolor{green!60!black}{\scriptsize (+0.9)}} \\
  & Win rate          & \textbf{64.0\%} & \textbf{63.5\%} & \textbf{71.7\%} & \textbf{71.3\%} & \textbf{67.6\%} \\
\midrule
\multirow{3}{*}{\text{Claude-Sonnet-4}}
  
  & \textsc{Base} SR  & 98.0\% & 97.3\% & 94.0\% & 74.7\% & 91.0\% \\
  & \textsc{Evolve} SR & \textbf{100.0\%}~{\textcolor{green!60!black}{\scriptsize (+2.0)}} & \textbf{99.3\%}~{\textcolor{green!60!black}{\scriptsize (+2.0)}} & \textbf{100.0\%}~{\textcolor{green!60!black}{\scriptsize (+6.0)}} & \textbf{100.0\%}~{\textcolor{green!60!black}{\scriptsize (+25.3)}} & \textbf{99.8\%}~{\textcolor{green!60!black}{\scriptsize (+8.8)}} \\
  & Win rate          & \textbf{66.0\%} & \textbf{56.8\%} & \textbf{68.1\%} & \textbf{56.2\%} & \textbf{61.8\%} \\
\midrule
\rowcolor{black!6}
\multicolumn{7}{l}{\textbf{(b) General prompt-following benchmarks: quality score (execution success \%)}} \\
\textbf{Backbone} & \textbf{Method}
 & \textbf{GenEval2}\,$\uparrow$ & \textbf{DPG-Bench}\,$\uparrow$
 & \textbf{OneIG-EN}\,$\uparrow$ & \textbf{OneIG-ZH}\,$\uparrow$ & \textbf{Avg}\,$\uparrow$ \\
\midrule
\multirow{3}{*}{\text{Claude-Opus-4.6}}
  & \textsc{Base}   &\textbf{61.24} \scriptsize(96.7) & 77.58 \scriptsize\textbf{(89.3)} & 60.66 \scriptsize(91.3) & 68.67 \scriptsize(93.3) & 67.04 \scriptsize(92.7) \\
  & \textsc{Evolve} & 57.92 \scriptsize\textbf{(98.0)} & \textbf{86.54} \scriptsize(84.7) & \textbf{68.79 \scriptsize(94.0)} & \textbf{70.21 \scriptsize(94.0)} & \textbf{70.87 \scriptsize(92.7)} \\
  & $\Delta$        & \textcolor{red}{-3.32} \scriptsize\textcolor{green!60!black}{(+1.3)} & \textcolor{green!60!black}{+8.96} \scriptsize\textcolor{red}{(-4.6)} & \textcolor{green!60!black}{+8.13} \scriptsize\textcolor{green!60!black}{(+2.7)} & \textcolor{green!60!black}{+1.54} \scriptsize\textcolor{green!60!black}{(+0.7)} & \textcolor{green!60!black}{+3.83} \scriptsize\textcolor{gray}{(0.0)} \\
\midrule
\multirow{3}{*}{\text{Claude-Sonnet-4}}
  & \textsc{Base}   & 34.26  \scriptsize(72.7) & 68.28 \scriptsize(82.7) & 52.00  \textbf{\scriptsize(98.0)} & 52.00  \scriptsize(96.7) & 51.63 \scriptsize(87.5) \\
  & \textsc{Evolve} & \textbf{46.25 \scriptsize(99.3)} & \textbf{83.79 \scriptsize(100)} & \textbf{53.47} \scriptsize(96.0) & \textbf{53.79} \scriptsize(96.7) & \textbf{59.33 \scriptsize(98.0)} \\
  & $\Delta$        & \textcolor{green!60!black}{+11.99} \scriptsize\textcolor{green!60!black}{(+26.6)} & \textcolor{green!60!black}{+15.51} \scriptsize\textcolor{green!60!black}{(+17.3)} & \textcolor{green!60!black}{+1.47} \scriptsize\textcolor{red}{(-2.0)} & \textcolor{green!60!black}{+1.79} \scriptsize\textcolor{gray}{(0.0)} & \textcolor{green!60!black}{+7.70} \scriptsize\textcolor{green!60!black}{(+10.5)} \\
\bottomrule
\end{tabular*}
\end{table*}

We evaluate whether evolving procedural memory from user-derived briefs improves design fulfillment across model scales: (i)~multi-round skill-bank dynamics (\S\ref{sec:evolution-dynamics}), (ii)~performance on external design benchmarks and held-out product briefs (\S\ref{subsec:main-results}), (iii)~transfer to general prompt following (\S\ref{subsec:general-t2i}), and (iv)~qualitative tool executions (\S\ref{sec:qualitative}). Section~\ref{sec:ablation} isolates widening and deepening.

\paragraph{Agents and backbones.}
\textsc{Evolve} equips the agent with the round-5 skill bank ($139$ skills), while \textsc{Base} disables skill access on the identical runtime and toolset. We evaluate \texttt{Claude-Opus-4.6}~\citep{anthropic2026claudeopus46} ($2{,}000$-token thinking budget) and \texttt{Claude-Sonnet-4}~\citep{anthropic2025claude4} ($5{,}000$-token budget) via Amazon Bedrock under a $900$\,s per-brief wall-clock cap.

\paragraph{Evolution traffic and held-out design briefs.}
Each round processes $\sim\!300$ briefs specifying target artifacts, text/layout constraints, and stylistic goals from historical user traffic and LLM-augmented variants ($1{,}406$ non-overlapping briefs and $1{,}869$ graded trajectories across five rounds). After each round, we evaluate \texttt{Claude-Sonnet-4} on $200$ disjoint human-authored briefs via \code{Graphic-Eval} (Appendix~\ref{sec:eval}), which scores requirement completeness in $[0,1]$ ($0$--$100$ in tables) alongside an independent visual aesthetics score.

\paragraph{External graphic-design benchmarks.}
We sample $300$ prompts each from OpenCOLE~\citep{Inoue_2024_CVPR}, GraphicBench~\citep{ki2025graphicbenchplanningbenchmarkgraphic}, CreatiDesign~\citep{zhang2026creatidesign}, and BannerRequest400~\citep{wang-etal-2025-banneragency}, reporting blind \textsc{Evolve}-vs.-\textsc{Base} win rates judged by \texttt{GPT-5.4}~\citep{openai2026gpt54thinking} under swapped orders alongside execution success.

\subsection{Skill-Bank Evolution Dynamics}
\label{sec:evolution-dynamics}

\paragraph{Selective admission and genuine coverage expansion.}
Across five rounds, the bank grows from $76$ seed skills to $139$ ($+63$ net): the replay gate commits $69$ widening additions and $131$ deepening revisions, rejects $67$ and $100$ proposals, and retires $6$ superseded or merged skills (Figure~\ref{fig:skill-flow}). Rejecting nearly half of all proposals confirms that matched replay actively filters noisy edits. In embedding space (Figure~\ref{fig:skill-umap}), newly acquired skill descriptions lie significantly farther from the seed bank than the seed bank's nearest-neighbor spacing (median cosine distance $0.215$ vs.\ $0.158$; Mann--Whitney $p<10^{-8}$, Cliff's $\delta=0.58$). Traffic-driven widening thus uncovers workflows absent from product documentation.

\paragraph{Shift from expansion to consolidation.}
Because widening waits for recurring capability gaps, skill additions surge in R2--R3 ($22$ and $26$ committed) once common uncovered subtasks surface. As those newly acquired V1 skills get retrieved in later traffic, evolution naturally shifts toward deepening: by R5, the gate commits $21$ revisions against only $4$ additions as the expanded bank stabilizes.

\paragraph{Local admission vs.\ whole-bank distribution.}
Per-round survival curves (Figure~\ref{fig:evolution-overview}) show why whole-bank held-out tracking complements local replay. Through R1--R4, new acquisitions steadily lift the high-completeness tail ($\tau\ge 0.9$ reaches $50\%$ by R4), yet R4 dips temporarily at lower thresholds ($\tau\ge 0.3$: $90\%$ vs.\ $94\%$ for \textsc{Base}) because rough V1 skills are only replay-tested on their triggering clusters and can misfire when retrieved on adjacent briefs. R5's revision pass resolves this cross-brief interference: retention reaches $98\%$ at $\tau\ge 0.3$ and exceeds \textsc{Base} at every threshold.

\subsection{Main Results on Design-Brief Fulfillment}
\label{subsec:main-results}

% ============================================================================
% Qualitative comparison grid — absolute-positioned TikZ.
% Needs from main.tex: graphicx, tikz, xcolor + the cc* colour tokens.
%
% IMAGES MUST BE SQUARE (1:1). \includegraphics only sets width, but the
% border is drawn at \imgW x \imgH — a non-square image will not line up
% with its frame. Run the square-crop pass before swapping in real results.
%
% Image filenames are explicit at each call so submission scanners can find them.
% ============================================================================
\begin{figure*}[t]
\centering
\setlength{\fboxsep}{0pt}%
\resizebox{\linewidth}{!}{%
\begin{tikzpicture}[
    font=\sffamily,
    inner sep=0pt, outer sep=0pt,
    every node/.style={inner sep=0pt, outer sep=0pt},
]
%% ---- Layout constants (cm) -------------------------------------------------
\def\imgW{1.58}          % image cell width  = square
\def\imgH{1.58}          % image cell height = square
% Column x-origins (pitch = imgW + 0.12 gap = 1.70)
\def\xA{0.55}\def\xB{2.25}\def\xC{3.95}
\def\xD{5.65}\def\xE{7.35}\def\xF{9.05}
\def\xR{10.73}           % right edge = xF + imgW + 0.10
% Row y-origins (pitch = imgH + 0.10 gap = 1.68)
\def\yTop{0.08}
\def\yHB{-1.30}          % header band bottom
\def\yBase{-1.40}        % Base row top
\def\yOurs{-3.08}        % Ours row top
\def\yBot{-4.76}         % figure bottom

%% ---- Background fills ------------------------------------------------------
\fill[white]      (-0.08,\yTop) rectangle (\xR+0.08,\yBot-0.06);
\fill[ccHeader]   (-0.08,\yTop) rectangle (\xR+0.08,\yHB);
\fill[ccOursTint] (-0.08,\yOurs+0.05) rectangle (\xR+0.08,\yOurs-\imgH-0.05);

%% ---- Horizontal rules ------------------------------------------------------
\draw[ccRule,   line width=0.50pt] (-0.08,\yHB) -- (\xR+0.08,\yHB);
\draw[ccCellBd, line width=0.28pt]
  (-0.08,\yBase-\imgH-0.05) -- (\xR+0.08,\yBase-\imgH-0.05);

%% ---- Macros ----------------------------------------------------------------
% #1 = column x-origin, #2 = prompt text
\def\casehead#1#2{%
  \fill[ccCell, rounded corners=1.5pt] (#1,-0.07) rectangle ++(\imgW,-1.16);%
  \draw[ccCellBd, line width=0.22pt, rounded corners=1.5pt]
    (#1,-0.07) rectangle ++(\imgW,-1.16);%
  \node[anchor=north, text width=1.46cm, align=center,
        font=\sffamily\fontsize{4.9pt}{6.0pt}\selectfont, text=ccText]
    at (#1+0.5*\imgW,-0.18) {#2};%
}
% #1 = row y-origin, #2 = label, #3 = colour
\def\methodlabel#1#2#3{%
  \node[font=\sffamily\bfseries\fontsize{6.2pt}{7.2pt}\selectfont, text=#3]
    at (0.27,#1-0.5*\imgH) {\rotatebox{90}{#2}};%
}
% #1 = x, #2 = y, #3 = image content
\def\imgcell#1#2#3{%
  \node[anchor=north west] at (#1,#2)
    {#3};%
  \draw[ccCellBd, line width=0.25pt] (#1,#2) rectangle ++(\imgW,-\imgH);%
}
\def\ourscell#1#2#3{%
  \node[anchor=north west] at (#1,#2)
    {#3};%
  \draw[ccOursGreen, line width=0.80pt] (#1,#2) rectangle ++(\imgW,-\imgH);%
}

%% ---- Prompt header row -----------------------------------------------------
\node[font=\sffamily\bfseries\fontsize{5.5pt}{6.5pt}\selectfont,
      text=ccTextDim, rotate=90] at (0.27,-0.65) {Prompt};

\casehead{\xA}{Create Adobe logo with double exposure effect of flowers}
\casehead{\xB}{Make a logo for Indian Coffee House}
\casehead{\xC}{A dog on the beach and add an eagle}
\casehead{\xD}{Soccer player's silhouette on the grass field}
\casehead{\xE}{Boy and girl on glass walkway, whale shark beneath}
\casehead{\xF}{Man with casual weekend attire on the sunny beach background}

%% ---- Row labels ------------------------------------------------------------
\methodlabel{\yBase}{Base}{ccText}
\methodlabel{\yOurs}{Evolve}{ccOursGreen}

%% ---- Base row --------------------------------------------------------------
\imgcell{\xA}{\yBase}{\includegraphics[width=\imgW cm]{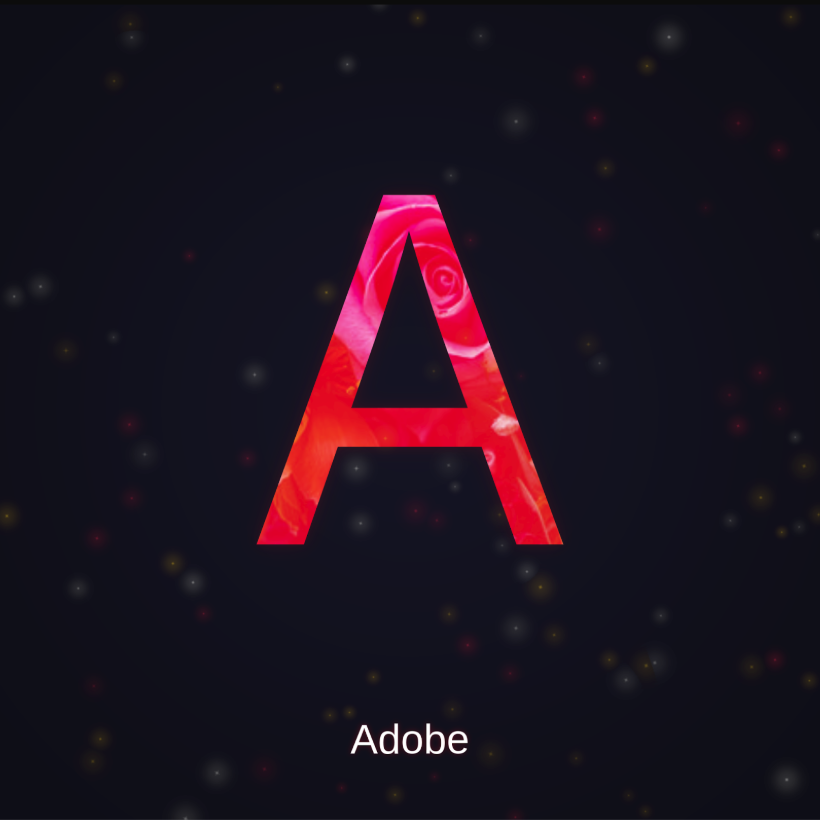}}
\imgcell{\xB}{\yBase}{\includegraphics[width=\imgW cm]{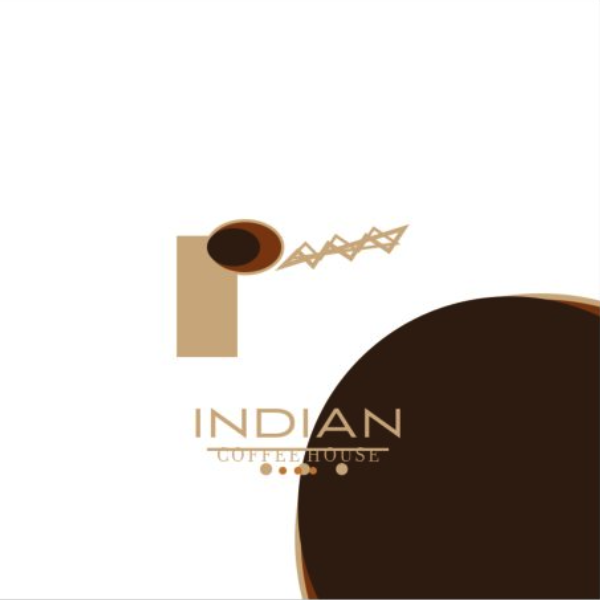}}
\imgcell{\xC}{\yBase}{\includegraphics[width=\imgW cm]{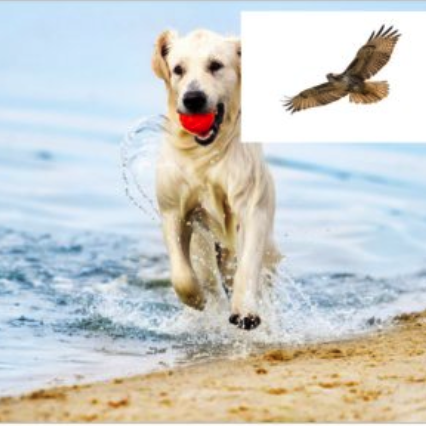}}
\imgcell{\xD}{\yBase}{\includegraphics[width=\imgW cm]{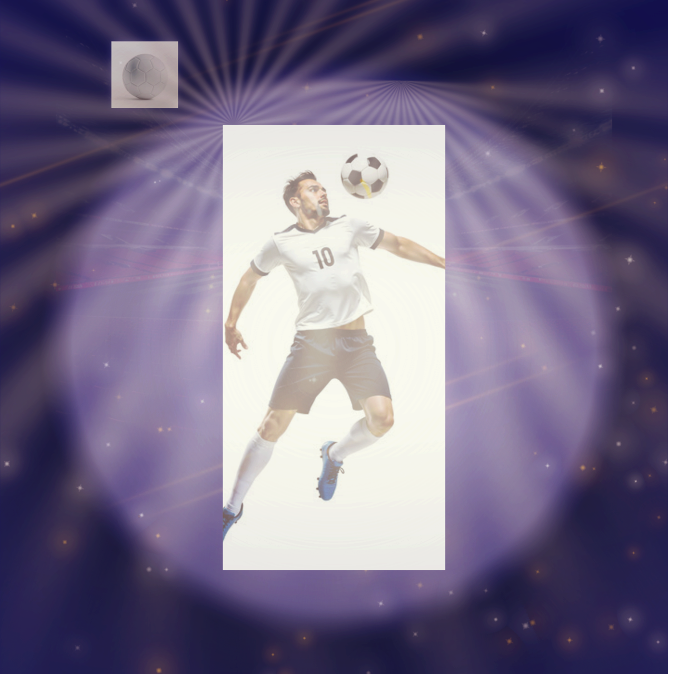}}
\imgcell{\xE}{\yBase}{\includegraphics[width=\imgW cm]{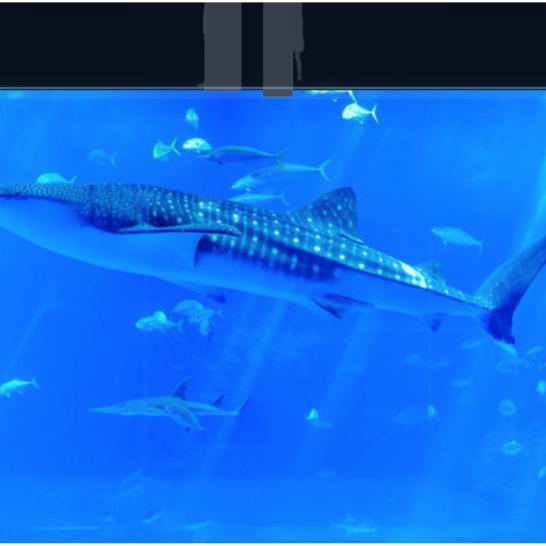}}
\imgcell{\xF}{\yBase}{\includegraphics[width=\imgW cm]{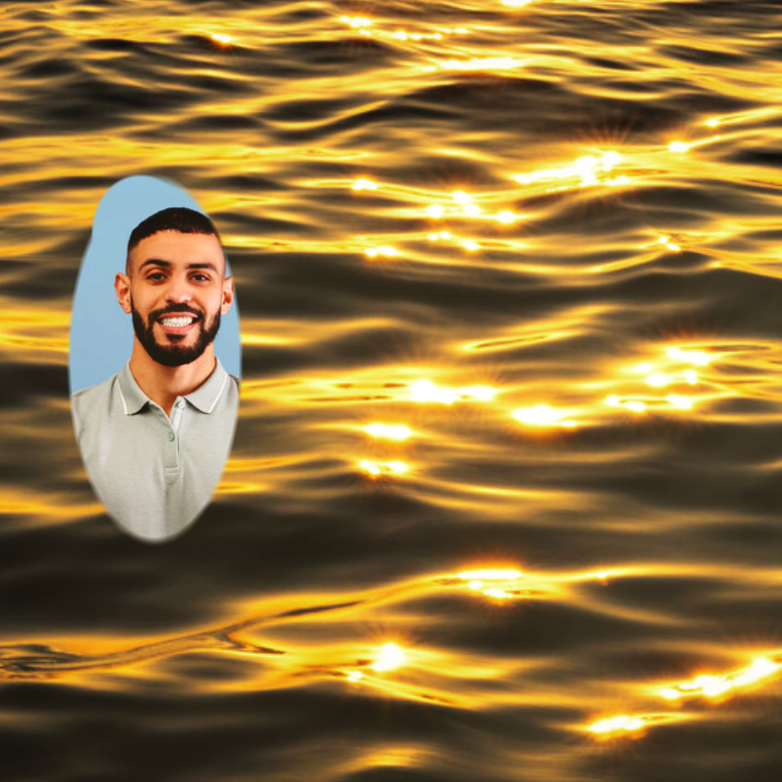}}

%% ---- Ours row --------------------------------------------------------------
\ourscell{\xA}{\yOurs}{\includegraphics[width=\imgW cm]{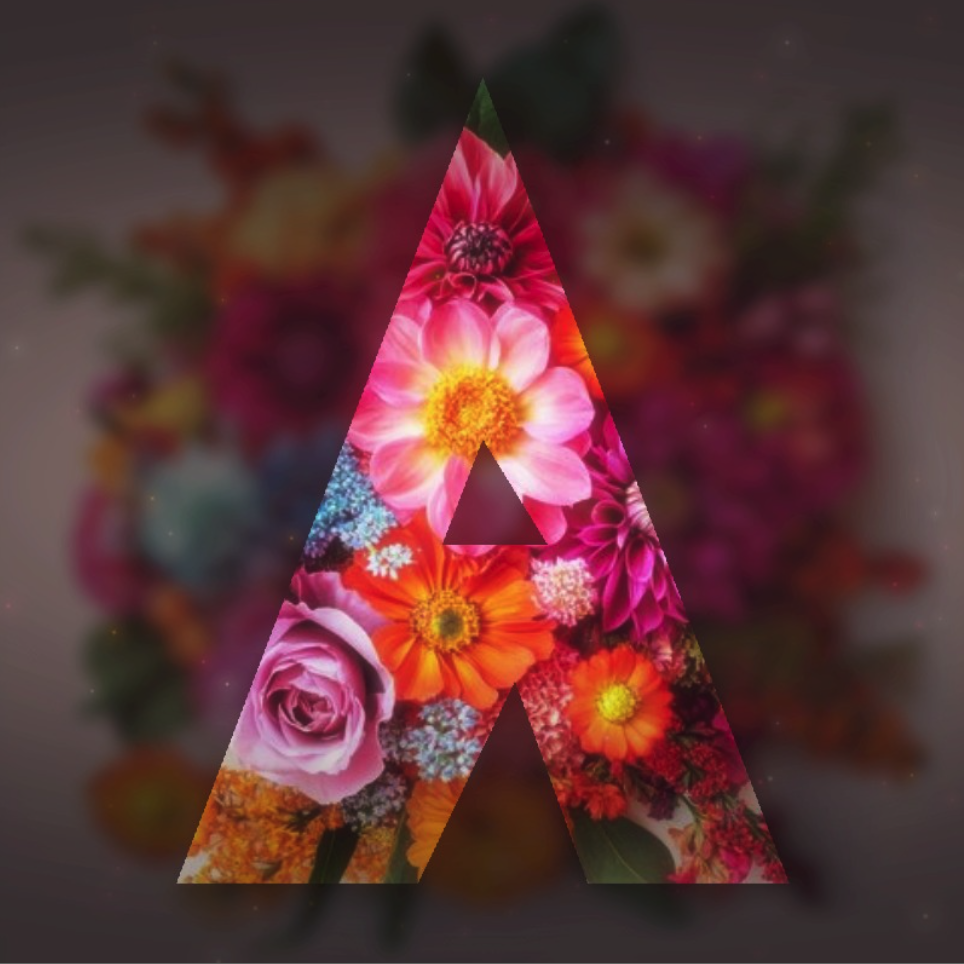}}
\ourscell{\xB}{\yOurs}{\includegraphics[width=\imgW cm]{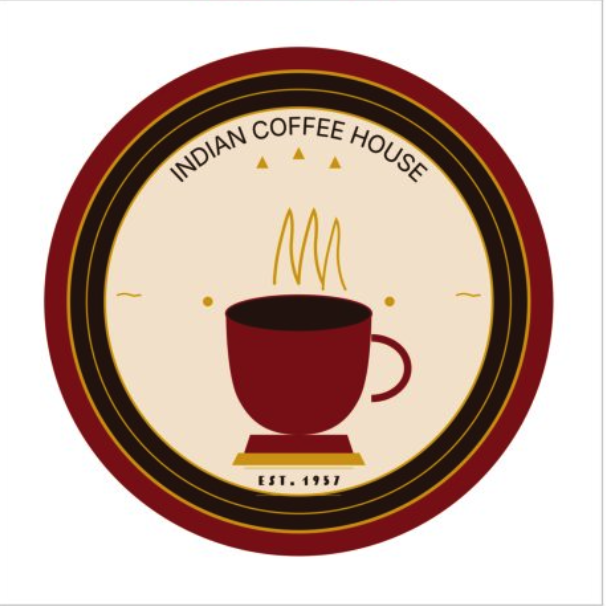}}
\ourscell{\xC}{\yOurs}{\includegraphics[width=\imgW cm]{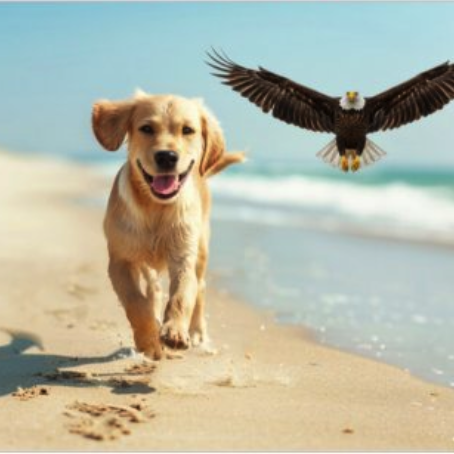}}
\ourscell{\xD}{\yOurs}{\includegraphics[width=\imgW cm]{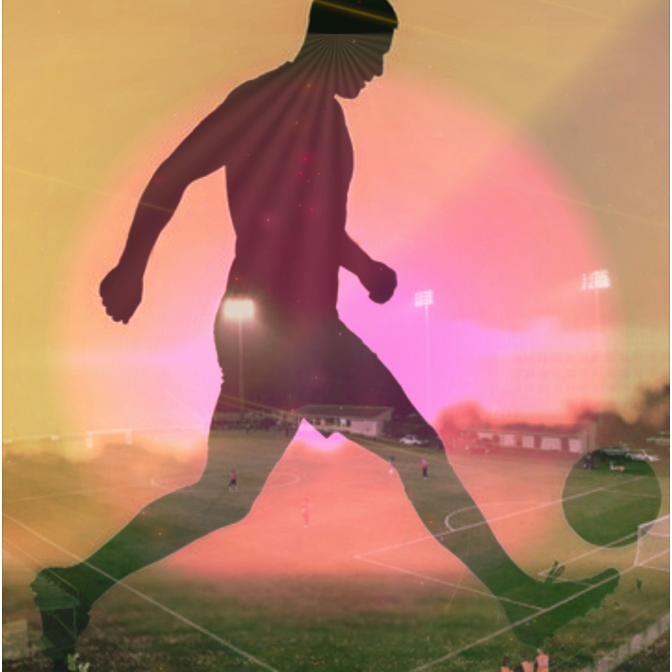}}
\ourscell{\xE}{\yOurs}{\includegraphics[width=\imgW cm]{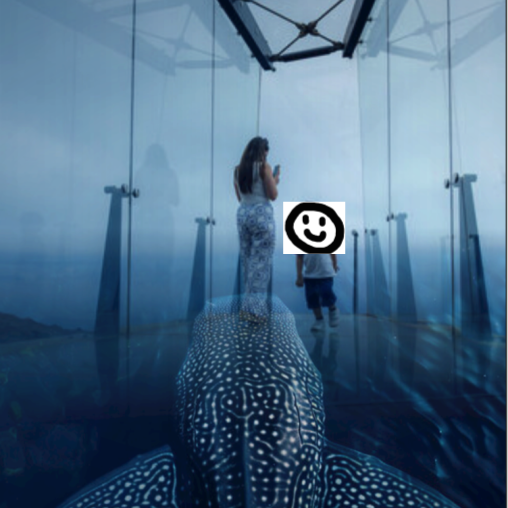}}
\ourscell{\xF}{\yOurs}{\includegraphics[width=\imgW cm]{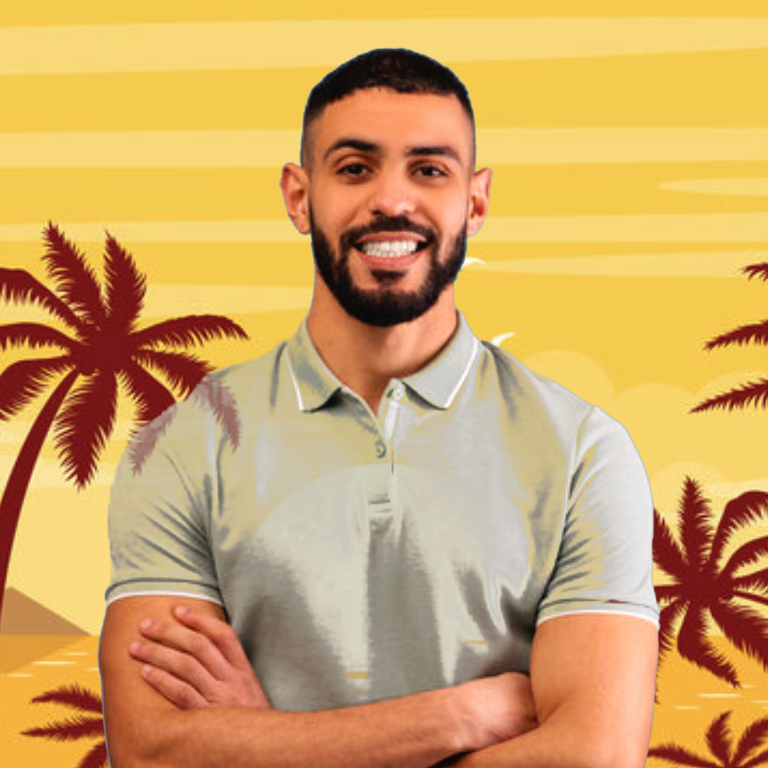}}

%% ---- Outer frame -----------------------------------------------------------
\draw[ccOuterBd, rounded corners=2.5pt, line width=0.45pt]
  (-0.08,\yTop+0.06) rectangle (\xR+0.08,\yBot-0.04);
\end{tikzpicture}%
}
\caption{\textbf{Qualitative comparison across six design briefs.} Columns correspond to distinct briefs, and rows compare \textsc{Base} (no skills) and \textsc{Evolve} (ours). Asset retrieval runs once per brief so both arms choose from the exact same candidate pool, isolating differences in asset selection and multi-step tool execution. All outputs are uncurated one-shot \texttt{Claude-Opus-4.6} executions (low thinking budget, no re-sampling or human editing), center-cropped to square for display.}
\label{fig:qual}
\end{figure*}

\paragraph{Consistent gains across benchmarks and model scales.}
\textsc{Evolve} beats \textsc{Base} on all four external design benchmarks for both backbones (Table~\ref{tab:main-results}(a)), achieving overall win rates of $61.8\%$ on \texttt{Claude-Sonnet-4} and $67.6\%$ on \texttt{Claude-Opus-4.6}, peaking at $68.1\%$ and $71.7\%$ on CreatiDesign. The stronger \texttt{Opus-4.6} backbone achieves even wider win-rate margins than \texttt{Sonnet-4}: procedural memory does not become redundant as model capability grows, because a stronger model follows complex multi-tool workflows more faithfully. On the smaller \texttt{Sonnet-4} model, evolved skills also substantially improve execution reliability, lifting average completion from $91.0\%$ to $99.8\%$ (including $74.7\%\rightarrow 100.0\%$ on BannerRequest400), while \texttt{Opus-4.6} rises from $98.3\%$ to $99.2\%$.

\paragraph{Higher requirement retention on held-out briefs.}
Evolved skills shift held-out completeness squarely toward full requirement fulfillment (Figure~\ref{fig:retention}). Relative to \textsc{Base}, the round-5 bank raises the share of briefs reaching $\tau\ge 0.5$ from $86\%$ to $93\%$, $\tau\ge 0.9$ from $43\%$ to $56\%$ ($+13$ points), and full completion ($\tau=1.0$) from $24\%$ to $32\%$.

\subsection{Transfer to General Prompt Following}
\label{subsec:general-t2i}

To test whether design-evolved procedures generalize beyond layout-heavy briefs, we evaluate four compositional and bilingual text-to-image suites under the same sampling protocol: GenEval2~\citep{kamath2025geneval2addressingbenchmark} (Soft-TIFA geometric mean), DPG-Bench~\citep{hu2024ellaequipdiffusionmodels} (Soft-TIFA arithmetic mean), and OneIG-EN/ZH~\citep{chang2025oneig} (VQAScore~\citep{Lin2024EvaluatingTG}), scored with \texttt{Qwen3-VL-8B-Instruct}~\citep{qwen3vltechnicalreport} over valid outputs. \textsc{Evolve} raises average quality by $+3.83$ points on \texttt{Opus-4.6} (improving three of four suites) and by $+7.70$ on \texttt{Sonnet-4} across all four suites (Table~\ref{tab:main-results}(b)). On \texttt{Sonnet-4}, evolved skills also lift average completion from $87.5\%$ to $98.0\%$, nearly eliminating execution failures on dense compositional prompts ($72.7\%\rightarrow 99.3\%$ on GenEval2 and $82.7\%\rightarrow 100\%$ on DPG-Bench).

\subsection{Qualitative Analysis of Tool Executions}
\label{sec:qualitative}

We compare one-shot \textsc{Base} and \textsc{Evolve} outputs when both arms receive the same retrieved stock-asset pool (Figure~\ref{fig:qual}), isolating asset choice and tool execution.

\paragraph{Executing multi-step masking and layout.}
Without procedural guidance, \textsc{Base} often stops at surface placement: it renders a plain red ``A'' omitting the flowers (Col.~1), leaves floating text over a dark box (Col.~2), pastes an unsegmented rectangular eagle patch onto the sky (Col.~3), and frames the soccer photo inside a box (Col.~4). Equipped with evolved skills, \textsc{Evolve} constructs letterform and subject alpha masks to blend floral and pitch textures through silhouettes (Cols.~1, 4), builds a circular emblem with curved banner typography (Col.~2), and composites a cleanly extracted eagle into the beach scene (Col.~3).

\paragraph{Choosing edit-compatible upstream assets.}
Skills also guide asset selection before editing begins. For the glass-walkway brief (Col.~5), \textsc{Base} picks an aquarium shot lacking the requested walkway, whereas \textsc{Evolve} selects an overhead tunnel with two figures and composites the whale shark beneath them. Similarly, for the beach portrait (Col.~6), \textsc{Base} produces garbled horizontal bands during foreground extraction, while \textsc{Evolve} selects an intact portrait and composites it onto a sunlit shore. Appendix~\ref{app:gallery} shows an uncurated $193$-prompt gallery across backbones.

\section{Ablation}
\label{sec:ablation}

\begin{table*}[t]
\centering
\caption{\textbf{Ablation of skill acquisition and revision on $200$ held-out briefs.} All variants run the same frozen agent in a single evaluation batch. Retaining only documentation skills (\textsc{Cold Start}), only revisions (\textit{+ rewrite}), or only unrevised V1 acquisitions (\textit{+ new skills}) fails to beat \textsc{Base}, whereas joint evolution (\textsc{Evolve}) yields super-additive gains in completeness and win rate. WR denotes pairwise win rate (\%) vs.\ \textsc{Base}, and SR denotes execution success (\%). Best values in \textbf{bold}.}
\label{tab:ablation}
\setlength{\tabcolsep}{4.5pt}
\renewcommand{\arraystretch}{1.15}
\begin{tabular*}{\linewidth}{@{}l@{\extracolsep{\fill}}rrrrrrrrrr@{}}
\toprule
& & \multicolumn{2}{c}{Quality $\uparrow$} & \multicolumn{2}{c}{Critique $\downarrow$} & \multicolumn{2}{c}{Tokens $\downarrow$} & & & \\
\cmidrule(lr){3-4}\cmidrule(lr){5-6}\cmidrule(lr){7-8}
Skill bank & \#Sk. & Compl. & Aesth. & Major & Minor & Prompt & Out & Tools & WR & SR \\
\midrule
\textsc{Base} & 0   & 69.08 & 65.92 & 3.93 & 8.02           & \textbf{346.8k} & \textbf{4979} & 33.5 & -----   & 95.0 \\
\midrule
\textsc{Cold Start}      & 76  & 68.62 & 65.66 & 3.88 & \textbf{7.87} & 445.2k          & 5298          & 35.4 & 46.4  & 96.5 \\
+ rewrite          & 76  & 69.02 & 65.98 & 3.70 & 8.14           & 450.1k          & 5369          & 35.9 & 48.6  & 97.5 \\
+ new skills       & 139 & 69.79 & 64.77 & 3.70 & 7.93           & 445.4k          & 5342          & 36.4 & 49.4  & \textbf{98.5} \\
\textsc{Evolve}   & 139 & \textbf{74.04} & \textbf{66.53} & \textbf{3.57} & 8.13 & 436.6k & 5167 & 36.7 & \textbf{58.5} & 98.0 \\
\bottomrule
\end{tabular*}
\end{table*}

We isolate our two update mechanisms on the held-out set in a single evaluation batch (Table~\ref{tab:ablation}, \texttt{GPT-5.4} and \code{Graphic-Eval}; Appendix~\ref{sec:eval}): \textit{+ rewrite} retains only the $76$ documentation-seeded skills at their final revised state (deepening only), whereas \textit{+ new skills} pairs the unrevised seed bank with the $63$ retained traffic-acquired skills at their initial V1 versions (widening only).

\paragraph{Static documentation skills do not help on their own.}
\textsc{Cold Start} slightly trails \textsc{Base} in completeness ($68.62$ vs.\ $69.08$) and wins only $46.4\%$ of comparisons despite adding $\sim\!98$k prompt tokens: product help pages document individual tool options, leaving multi-step composition and error recovery unaddressed. Traffic-driven evolution (\textsc{Evolve}) reverses this deficit, reaching $74.04$ completeness ($+4.96$ over \textsc{Base}) and a $58.5\%$ win rate ($p=0.025$).

\paragraph{Synergy between widening and deepening.}
Neither operation succeeds in isolation: relative to \textsc{Cold Start}, deepening alone (\textit{+ rewrite}) adds $+0.40$ completeness ($48.6\%$ WR) and widening alone (\textit{+ new skills}) adds $+1.17$ ($49.4\%$ WR). Combining both in \textsc{Evolve} yields $+5.42$, which exceeds the sum of the two standalone gains ($+1.57$) by \textbf{$3.85$ points}. Revising seed skills repairs flawed procedures but cannot address missing capabilities, whereas unrevised V1 skills broaden task completion ($98.5\%$ SR) with rough first-draft playbooks distilled from only a few trajectories, and aesthetics drops to $64.77$. Subsequent deepening turns that raw coverage into reliable execution.

\paragraph{Fulfillment gains at bounded inference cost.}
Gains concentrate in requirement fulfillment alongside modest improvements in aesthetics ($66.53$ vs.\ $65.92$) and major-defect counts ($3.57$ vs.\ $3.93$). Despite nearly doubling the indexed skill catalog over \textsc{Cold Start}, \textsc{Evolve} consumes slightly fewer prompt and output tokens per brief ($436.6$k vs.\ $445.2$k prompt tokens, cached context included), because two-stage retrieval and on-demand \code{load\_skill} disclosure inject only the procedures invoked for the active brief.

\section{Limitations and Future Work}

Currently, Designer-RSI updates a shared skill bank in offline batches, which trades replay compute and VLM grading overhead for reliability on single-turn, general-purpose briefs. A natural next step is moving to multi-turn design sessions, where user goals, document state, and feedback shift as the canvas takes shape. In that setting, sustained human--agent collaboration means learning reusable procedures directly from user edits and demonstrations, while knowing when to step back and let the designer take over. Beyond shared workflows, we also want the skill bank to adapt to individual users, retrieving and revising skills around a designer's preferred visual style and construction habits as they evolve over time. Building on our preliminary user-guided study (Appendix~\ref{app:personalization}), we plan to evaluate how well personalized skills reduce human effort and improve alignment across repeated sessions.

\section{Conclusion}

We presented Designer-RSI and showed that recursive self-improvement can operate reliably in complex creative software without ground-truth verifiers, human annotations, or weight updates. Three broader lessons emerge for self-evolving multimodal agents. First, executable procedural memory cannot be distilled from static product manuals alone: reliable skills must encode cross-tool sequencing, asset selection, and failure recovery grounded in real execution traces. Second, expanding coverage (\textbf{widening}) and refining execution (\textbf{deepening}) depend on each other, as newly induced skills require contrastive trajectory-level revision before they become dependable. Third, separating open-ended skill proposal from conservative admission via a pairwise \textbf{matched replay gate} curbs regressions under noisy VLM judging and stochastic asset retrieval. Together, evolving modular, human-auditable procedural memory from user traffic gives industrial product agents a practical path to continual self-improvement in production.

\setcounter{topnumber}{2}
\setcounter{dbltopnumber}{2}

% Mandatory LLM Usage Statement (allowed on the reference pages per CVPR 2027 author guidelines)

% \input{sections/ai_usage}

{
    \small
    \bibliographystyle{assets/plainnat}
    \bibliography{references}
}

% WARNING: the supplementary below must be split off into a separate PDF before submission.

% ===== Supplementary material (CVPR kit workflow) =====
% Compiled with the main paper so cross-references resolve. split_pdf.py (run
% automatically by .latexmkrc after each build) splits main.pdf at this point into
% main_paper.pdf and supplementary.pdf, using the page count logged below.
\clearpage
\typeout{SUPP-SPLIT: main pages = \the\ReadonlyShipoutCounter}
\onecolumn
\setcounter{page}{1}
\begin{center}{\Large\bfseries \designerrsititle\par}\vspace{0.5em}{\large Supplementary Material\par}\end{center}
\vspace{1em}
\appendix

% ---- Appendix table of contents (clickable via hyperref) ----
\addtocontents{toc}{\protect\setcounter{tocdepth}{2}}
\begingroup
\renewcommand{\contentsname}{Appendix Contents}
\hypersetup{linkcolor=black}
% Spread the directory across its dedicated page with clear section spacing.
\fontsize{11}{18}\selectfont
\makeatletter
\let\originaltocsection\l@section
\renewcommand*\l@section[2]{%
  \ifnum\c@tocdepth>\z@
    \addvspace{20pt}%
  \fi
  \originaltocsection{#1}{#2}%
}
\makeatother
\tableofcontents
\endgroup
\newpage
\section{Additional Graphic Design Agent Details}
\label{app:agent_detail}

\paragraph{Runtime.}
Each user request invokes an iterative tool-calling loop in which the model
plans, executes tools, observes their structured outputs, and revises the design.
The runtime supports parallel tool execution, parameter validation and repair,
failure recovery, multimodal previews, and context compression for long
trajectories.

\paragraph{Asset-grounded creation and rendering.}
The agent constructs designs from retrieved assets rather than generating all
pixels directly. Candidate assets are retrieved from Adobe Stock
\citep{adobestock}, enriched with visual attributes such as dominant color,
subject position, and text-safe regions, and diversified using maximal marginal
relevance \citep{carbonell1998mmr}. A production render engine provides
typography using global paragraph optimization \citep{knuth1981breaking},
compositing, masks, effects, gradients, multilingual layout following the
Unicode bidirectional algorithm \citep{unicode2025bidi}, and subject
segmentation based on U$^2$-Net \citep{qin2020u2net}. Designs can be exported
to editable application formats as well as print-ready PDF.

\paragraph{Verification and offline evaluation.}
During execution, the agent periodically renders its current document for
multimodal inspection and runs deterministic checks over properties including
text overflow, occlusion, alignment, visual balance, and contrast, including
WCAG-style contrast constraints \citep{wcag22}. Document state is serializable,
allowing completed trajectories to be deterministically rendered and evaluated
offline. This separates the comparatively expensive multimodal reward
computation from the user-facing execution path.
\section{Skill Bank Architecture and Retrieval}
\label{sec:bank}

The skill bank is the agent's inspectable procedural memory: a versioned collection of natural-language \code{SKILL.md} files, each encoding a reusable multi-tool workflow (Section~\ref{sec:intro}). Here we detail its static representation, retrieval, progressive injection, and trajectory logging. Section~\ref{sec:evolve} describes how offline evolution updates it.

\subsection{Skill Representation}
Each skill is stored as a single Markdown file (\code{SKILL.md}) with YAML front matter inside a directory matching its \code{name}. The bank is partitioned into three sub-banks by \code{app\_mode}: \code{raster} (Photoshop-equivalent), \code{vector} (Illustrator-equivalent), and \code{page\_layout}/\code{any} (InDesign-equivalent and cross-application). The YAML front matter specifies:
\begin{itemize}
  \item \code{description}: $200$--$600$ characters of operation-centric trigger phrases. This field is the \textbf{sole retrieval signal}, while the body text affects only what the agent reads after loading the skill.
  \item \code{app\_mode} $\in\{$\code{raster}, \code{vector}, \code{page\_layout}, \code{any}$\}$: restricts scoring to compatible application sessions.
  \item \code{tool\_sequence}: the $5$--$10$-tool critical path defining the core workflow (and, optionally, a scoped tool whitelist).
  \item \code{prerequisites}: dependent skills automatically surfaced (one hop) alongside a match.
  \item \code{status} $\in\{$\code{candidate}, \code{stable}, \code{deprecated}$\}$ and an integer \code{version} counter.
\end{itemize}
Every skill body contains five required sections: \emph{when to use}, \emph{clarify before starting}, \emph{steps} (naming concrete tools and key parameters), \emph{tips} (non-obvious domain constraints), and \emph{error handling} (a situation-to-recovery table). Per-skill execution statistics are stored externally so each \code{SKILL.md} remains clean and human-editable.

\subsection{Two-Stage Lexical Retrieval}
\label{sec:retrieval}
Because user briefs are typically verbose and subject-heavy whereas skills are operation-centric, runtime retrieval proceeds in two stages.

\paragraph{(1) Query distillation.} A frozen LLM maps the user brief to at most six core design operations (e.g., ``double exposure, background removal''), explicitly stripping subject nouns (names, brands, locations), style adjectives, and color values. This aligns the query directly with the operation-centric \code{description} fields in the bank.

\paragraph{(2) Token-overlap ranking.} Each skill $s$ is scored by lexical token overlap between the distilled query $q$ and its description:
\begin{equation}
  \mathrm{score}(s) \;=\; \frac{\bigl|\,\mathrm{tokens}(q)\cap\mathrm{tokens}(\mathrm{desc}(s))\,\bigr|}{\bigl|\,\mathrm{tokens}(q)\,\bigr|}.
  \label{eq:retrieval}
\end{equation}
Each distilled operation is scored independently, with a name-match bonus favoring skills named for that operation over incidental token matches, and \code{app\_mode} zeros out incompatible skills (e.g., a \code{raster} skill scores $0$ in a \code{vector} session). Retrieval keeps the full bank accessible while flagging \textbf{$2$--$4$} skills per turn as top candidates: those matching an operation name or clearing $\tau=0.2$ within $0.6\times$ of the top score (up to $4$, padded to $2$ when fewer qualify).

\paragraph{Why lexical overlap over dense embeddings.} Using lexical overlap aligns cold-start documentation seeding (Appendix~\ref{sec:coldstart}) and runtime retrieval on the exact same tokenizer, making the relationship between a skill's authored trigger phrases and its routing behavior transparent and debuggable during deepening edits.

\subsection{Progressive Disclosure via \code{load\_skill}}
To avoid concatenating full skill bodies into every system prompt, the runtime uses progressive disclosure. Once per turn, an injector appends a compact catalog listing each skill's name, \code{app\_mode} tag, and one-line \code{description} (bounded by a token budget), with the top matches from Appendix~\ref{sec:retrieval} starred and floated to the top. When the model deems a skill relevant, it invokes \code{load\_skill(name)}, which returns that skill's full steps, tool sequence, tips, and error-handling table as a tool observation. Only chosen skills thus enter the active context. The catalog spans all three applications so cross-application workflows can be planned upfront, while active tool lists remain scoped to the current document's application. Optionally, the active toolset can be narrowed further to the union of loaded skills' \code{tool\_sequence} lists plus a small core set (disabled by default in our main experiments, and enabled where noted in Appendix~\ref{app:personalization}). Turns with no starred matches are logged as uncovered subtasks (Section~\ref{sec:evolve}).

\subsection{Trajectory Logging}
For every turn, the runtime records the distilled query, retrieved skills and overlap scores, executed tool calls, coverage tag, and a pointer to the serialized \code{XML} document snapshot. The offline grader populates outcome scores and requirement rationales asynchronously, supplying the traces needed for deepening failure attribution and widening gap clustering.

\section{Evaluation Rubric: Graphic-Eval}
\label{sec:eval}

\textbf{Graphic-Eval} provides offline evolution feedback and evaluates held-out internal briefs under a fixed rubric along three axes: requirement completeness, visual aesthetics, and diagnostic critique. External benchmark metrics and blind pairwise protocols are detailed in Section~\ref{sec:experiments}.

\subsection{Deterministic Rendering}
Grading operates directly on rendered images (not raw document trees). Each turn's serialized \code{XML} snapshot is rendered to PNG at a fixed resolution. Multi-turn sessions are evaluated against the cumulative brief (initial prompt plus all subsequent user refinements up to that turn), so the graded artifact must satisfy every accumulated constraint.

\subsection{Requirement Completeness (Decompose and Check)}
\label{sec:completeness}
Completeness measures how thoroughly the rendered output fulfills the brief using two structured LLM/VLM calls.

\paragraph{Decompose.} A text LLM call, prompted as a senior graphic designer, extracts $3$--$10$ verifiable requirements from the brief. Each requirement is phrased so it can be judged \code{met}, \code{partial}, or \code{not\_met}, preserving interpretive flexibility for stylistic goals (e.g., ``evoke a vintage mood'') while keeping literal text and layout constraints exact. Related attributes are grouped together, and tool- or software-specific prescriptions are explicitly \textbf{excluded} so completeness grades visual outcomes independently of the implementation path.

\paragraph{Check.} A multimodal VLM call receives the brief, the extracted requirement list, and the rendered PNG, labeling each requirement \code{met}, \code{partial}, or \code{not\_met} with an obligatory failure rationale whenever a requirement is not fully met. Completeness is the mean earned fraction:
\begin{equation}
  \mathrm{completeness} \;=\; \frac{1}{N}\sum_{i=1}^{N}
  \begin{cases}
    1.0 & \text{if requirement } i \text{ is \code{met}}\\
    0.5 & \text{if \code{partial}}\\
    0.0 & \text{if \code{not\_met}}
  \end{cases}
  \label{eq:completeness}
\end{equation}
The per-requirement failure rationales supply the diagnostic signal for the Reflector during deepening (Section~\ref{sec:evolve}).

\subsection{Visual Aesthetics (Anchored 1--10)}
A separate VLM call scores pure visual craft (composition, balance, negative space, typographic hierarchy, color harmony, form language, and alignment/readability) while explicitly ignoring brief adherence so aesthetics and completeness remain disentangled. The judge assigns an integer in $1$--$10$ against calibrated anchors ($6$ = passable, $8$ = polished professional work, $9$--$10$ = exceptional, $\le 3$ = poor) alongside a brief summary, stored normalized as $\mathrm{aesthetic}=\text{score}/10 \in [0,1]$. Decoupling aesthetics from prompt adherence motivates its smaller weight $w_a$ in Eq.~(\ref{eq:passrate}): a visually polished design that ignores the user's request cannot achieve a passing trajectory score.

\subsection{Diagnostic Critique (Defect Counts)}
A third multimodal call, prompted as a design director, audits eight craft dimensions (hierarchy/typography, contrast/readability, spacing/alignment, color, composition/balance, execution craft, consistency, and originality), listing visible defects tagged as \code{major} or \code{minor} while omitting clean dimensions and proposing no edits. Unlike scalar scores, major and minor defect counts localize specific visual weaknesses for ablation diagnostics (Table~\ref{tab:ablation}).

\subsection{Judge Backends and Composite Scoring}
Trajectory binning combines completeness and aesthetics via Eq.~(\ref{eq:passrate}) with $(w_c,w_a)=(0.7,0.3)$ (Appendix~\ref{app:scoring}). \code{Graphic-Eval} supports Azure-hosted GPT and Bedrock-hosted Claude backends under identical prompts and strict JSON schemas, held fixed throughout evolution. Update admission bypasses scalar scores and uses the position-balanced pairwise replay gate in Section~\ref{sec:gate}.

\subsection{Inference Telemetry}
Every trajectory logs tool calls, wall-clock latency, and prompt/output tokens (including cached context) from completion metadata. Table~\ref{tab:ablation} reports per-request inference token budgets, separate from one-time offline evolution compute.

\section{Trajectory Scoring and Candidate Triggering}
\label{app:scoring}

Every recorded trajectory $j$ is graded by frozen \code{Graphic-Eval} (Appendix~\ref{sec:eval}), yielding a completeness fraction, a normalized aesthetic score, and unmet-requirement rationales. For candidate selection, we combine both scalar axes with completeness prioritized:
\begin{equation}
  s_j \;=\; w_c \cdot \mathrm{completeness}_j \;+\; w_a \cdot \mathrm{aesthetic}_j,
  \qquad (w_c,w_a)=(0.7,\,0.3).
  \label{eq:passrate}
\end{equation}
Prioritizing completeness prevents visually appealing outputs that ignore prompt constraints from masking task failures.

\paragraph{Outcome binning.} Each trajectory is binarized at $\tau=0.6$: $s_j \ge \tau$ is labeled a success and $s_j < \tau$ a failure. Every skill retrieved during a failed trajectory increments its failure count with uniform credit assignment (no per-skill blame heuristic), and successful trajectories similarly log positive usage histories. Absolute VLM scores are used \emph{exclusively} in this binning step to flag revision candidates. No skill update is ever admitted from scalar score changes due to run-to-run VLM judge drift (Appendix~\ref{app:gate}).

\paragraph{Triggering revision candidates.} Once at least $50$ newly graded records accumulate (batched at $\sim\!300$ briefs per round in our offline experiments), skills with failure count $\ge m$ ($m=2$) are queued for deepening in descending order of failures, capped at $50$ skills per round. Relying solely on raw failure count, without suppressing high-traffic skills that also succeed frequently, ensures widely retrieved skills still undergo refinement when recurring edge cases appear. An exponential moving average of binary outcomes (success $\rightarrow 1$, failure $\rightarrow 0$, reset to $0.5$ upon each version update) is tracked for monitoring only and does not affect routing or selection.

\section{Matched Replay Gate Details}
\label{app:gate}

\paragraph{Why relative, matched-context evaluation.} Because scalar VLM grades vary across repeated evaluations of the same image and agent executions depend heavily on upstream asset choices, the gate evaluates every candidate update via freshly executed, head-to-head visual comparisons under identical upstream state and swapped presentation orders.

\paragraph{Stratified replay sets and baselines.} Replay sampling adapts to the update operation:
\begin{itemize}
  \item \textbf{Deepening (skill revision):} Replay briefs are drawn from historical trajectories that retrieved the incumbent skill, stratified by outcome: two from previously successful trajectories ($s_j \ge 0.6$) and two from failed ones ($s_j < 0.6$). The failed pair tests whether the candidate resolves the target failure modes, while the successful pair guards against regressions on cases the incumbent already handled. The candidate version (V2) is replayed directly against the incumbent version (V1).
  \item \textbf{Widening (skill acquisition):} Because a newly distilled skill has no prior version and targets a capability gap, replay briefs are sampled directly from the triggering gap cluster without score stratification, and the candidate skill is replayed against the no-skill \textsc{Base} agent.
\end{itemize}
Both modes run candidate and baseline arms in the same batch from identical retrieved assets and document state.

\paragraph{Pairwise judging and tie band.} For each sampled brief, candidate and baseline renders across matched contexts are compared by a pairwise VLM judge asking which image better fulfills the brief, evaluated in both presentation orders $(A,B)$ and $(B,A)$ to cancel position bias~\citep{wang2024fair,zheng2023judging}. Aggregating votes yields a per-brief win rate $r \in [0,1]$ for the candidate. A symmetric tie band of half-width $\delta$ around $0.5$ absorbs judge noise on near-identical renders: a brief counts as won if $r > 0.5+\delta$, lost if $r < 0.5-\delta$, and tied otherwise.

\paragraph{Acceptance rule.}
\begin{equation}
  \textbf{accept} \quad\Longleftrightarrow\quad
  \bigl(\nexists\text{ sampled brief with } r < 0.5-\delta\bigr)\ \wedge\
  \bigl(\exists\text{ sampled brief with } r > 0.5+\delta\bigr).
  \label{eq:accept-appendix}
\end{equation}
Candidates that merely tie across all sampled briefs are rejected, enforcing strict improvement on at least one brief with zero detected regressions, an empirical analogue of baseline-bootstrapped policy screening~\citep{laroche2019safe,thomas2015highconfidence}.

\paragraph{Handling rejections.} When a deepening revision is rejected, it is discarded and the skill's consecutive-rejection counter increments (triggering escalation from targeted edit to full rewrite, and subsequently one exploration cycle). When a widening candidate is rejected, the draft \code{SKILL.md} is discarded while its triggering occurrences remain in the coverage pool so later rounds can re-distill the skill once additional examples accumulate. Every accepted and rejected proposal is version-logged with before/after snapshots for auditability.

\section{Cold Start from Product Documentation}
\label{sec:coldstart}

Because a newly deployed agent has no execution traffic and an empty bank retrieves nothing, we initialize the skill bank (\textsc{Cold Start}) from official product help documentation via a four-phase pipeline before evolution begins. Phases invoking an LLM use the same frozen model as the rest of the framework, while two phases are purely deterministic:

\begin{description}
  \item[Phase 1: Intent clustering.] Stream English product help articles, deduplicate entries, and extract per-application topics. A small number of LLM calls group these topics into $20$--$25$ coherent workflow themes per application, each representing a candidate skill.
  \item[Phase 2: Lexical page assignment (no LLM).] Score every help page against each theme using the exact same lexical token-overlap tokenizer used at runtime (Appendix~\ref{sec:retrieval}) and attach the top-scoring pages to each theme, aligning authored descriptions with runtime retrieval.
  \item[Phase 3: Skill synthesis.] One LLM call per theme synthesizes a draft \code{SKILL.md} grounded in its assigned documentation pages, the live tool catalog, and existing skill descriptions for deduplication. A \code{NO\_SKILL} escape token filters out themes that cover only UI navigation without reusable design workflows.
  \item[Phase 4: Deterministic validation (no LLM).] Verify every referenced tool name against the live $230+$ tool catalog, reject near-duplicate descriptions, enforce all five required Markdown body sections, write valid files into their respective \code{app\_mode} sub-banks, and initialize neutral monitoring statistics.
\end{description}

Lightweight human sanity checks follow Phase 1 (theme inspection) and Phase 3 (draft inspection), producing the \textbf{$76$-skill \textsc{Cold Start} bank} ($26$ Photoshop-, $25$ Illustrator-, and $25$ InDesign-equivalent or cross-application skills). As shown in Section~\ref{sec:ablation}, documentation-derived skills alone fail to outperform \textsc{Base}. Traffic-driven evolution (Section~\ref{sec:evolve}) is required to adapt the bank to real multi-step briefs.

\section{Personalizing Skills to User-Specific Methods}
\label{app:personalization}

\begin{figure}[t]
\centering
\setlength{\tabcolsep}{0pt}
\begin{tabularx}{\linewidth}{@{}>{\raggedright\arraybackslash}p{0.28\linewidth}@{\hspace{4pt}}>{\centering\arraybackslash}X@{\hspace{4pt}}>{\centering\arraybackslash}X@{}}
{\centering\small\textbf{Prompt}\par} & \small\textbf{No skill} (\textsc{Base}) & \small\textbf{With skill} (Ours) \\[4pt]
\scriptsize\textbf{Method preference.}\ ``Add a snow effect to this night-street photo.'' &
  \adjustbox{valign=t}{\includegraphics[width=\linewidth]{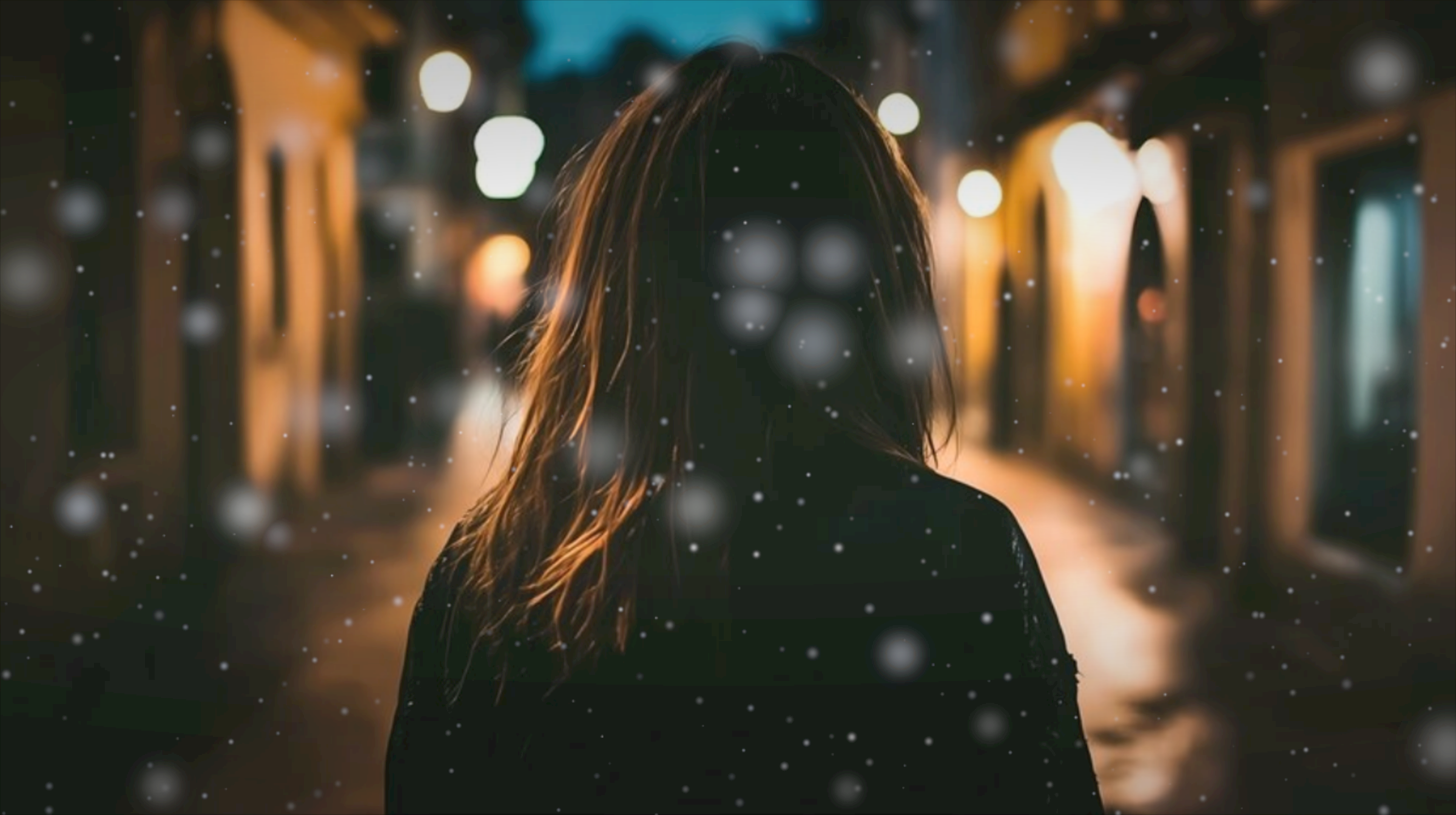}} &
  \adjustbox{valign=t}{\includegraphics[width=\linewidth]{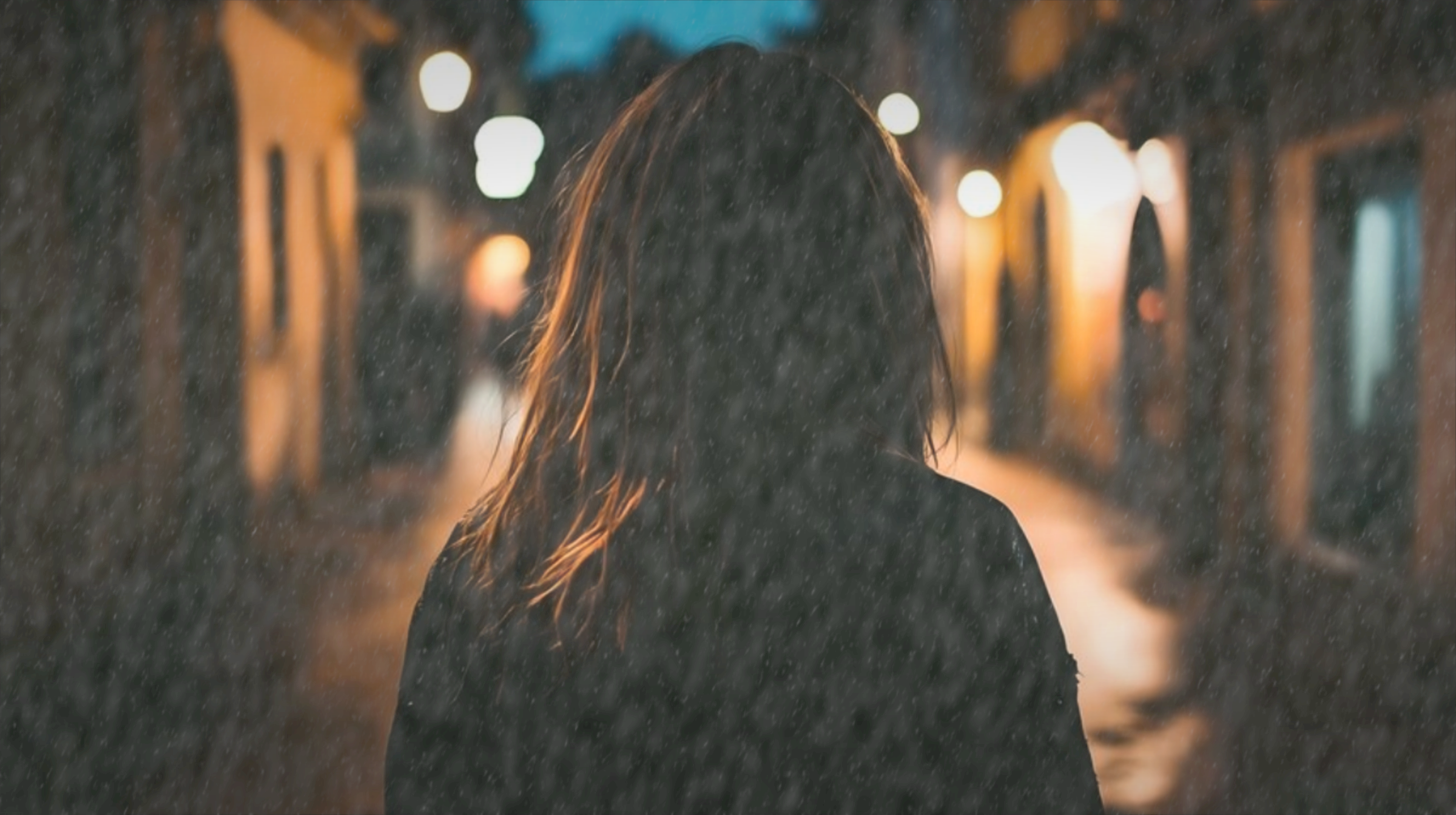}} \\
\noalign{\vskip 4pt}
\scriptsize\textbf{Perception / precision.}\ ``Make a three-layer Droste effect from this framed picture.'' &
  \adjustbox{valign=t}{\includegraphics[width=\linewidth]{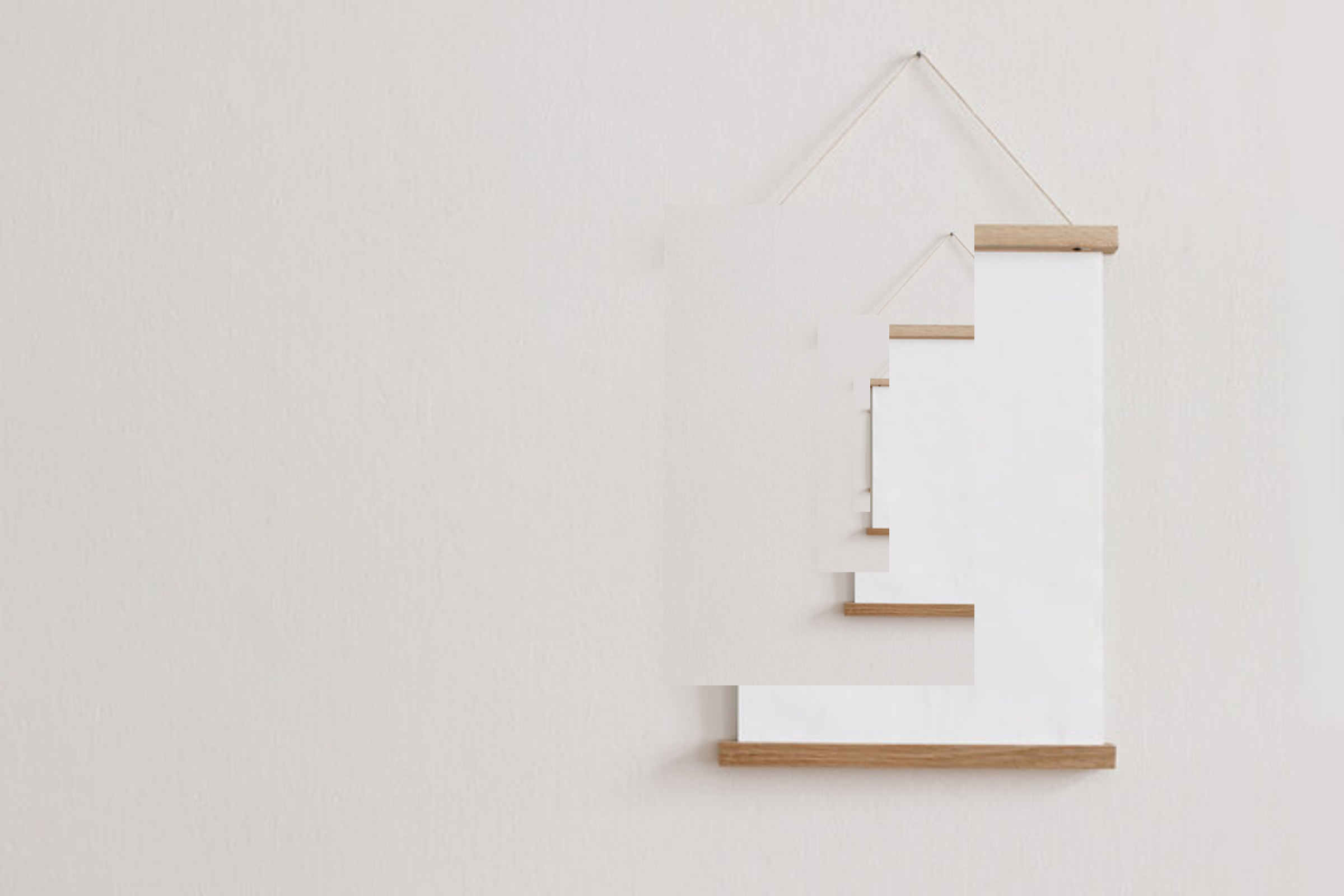}} &
  \adjustbox{valign=t}{\includegraphics[width=\linewidth]{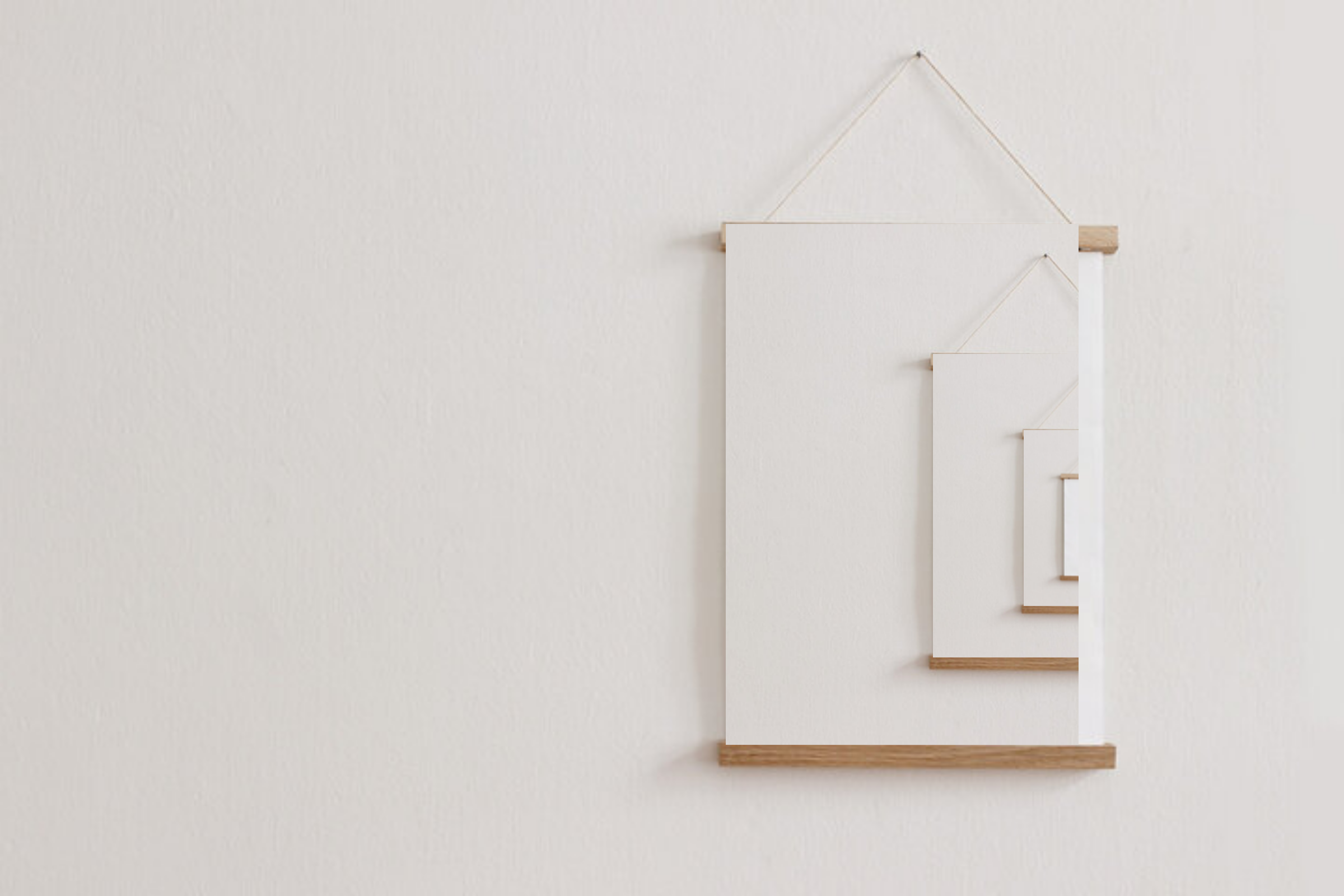}} \\
\noalign{\vskip 4pt}
\adjustbox{valign=t}{\includegraphics[width=\linewidth]{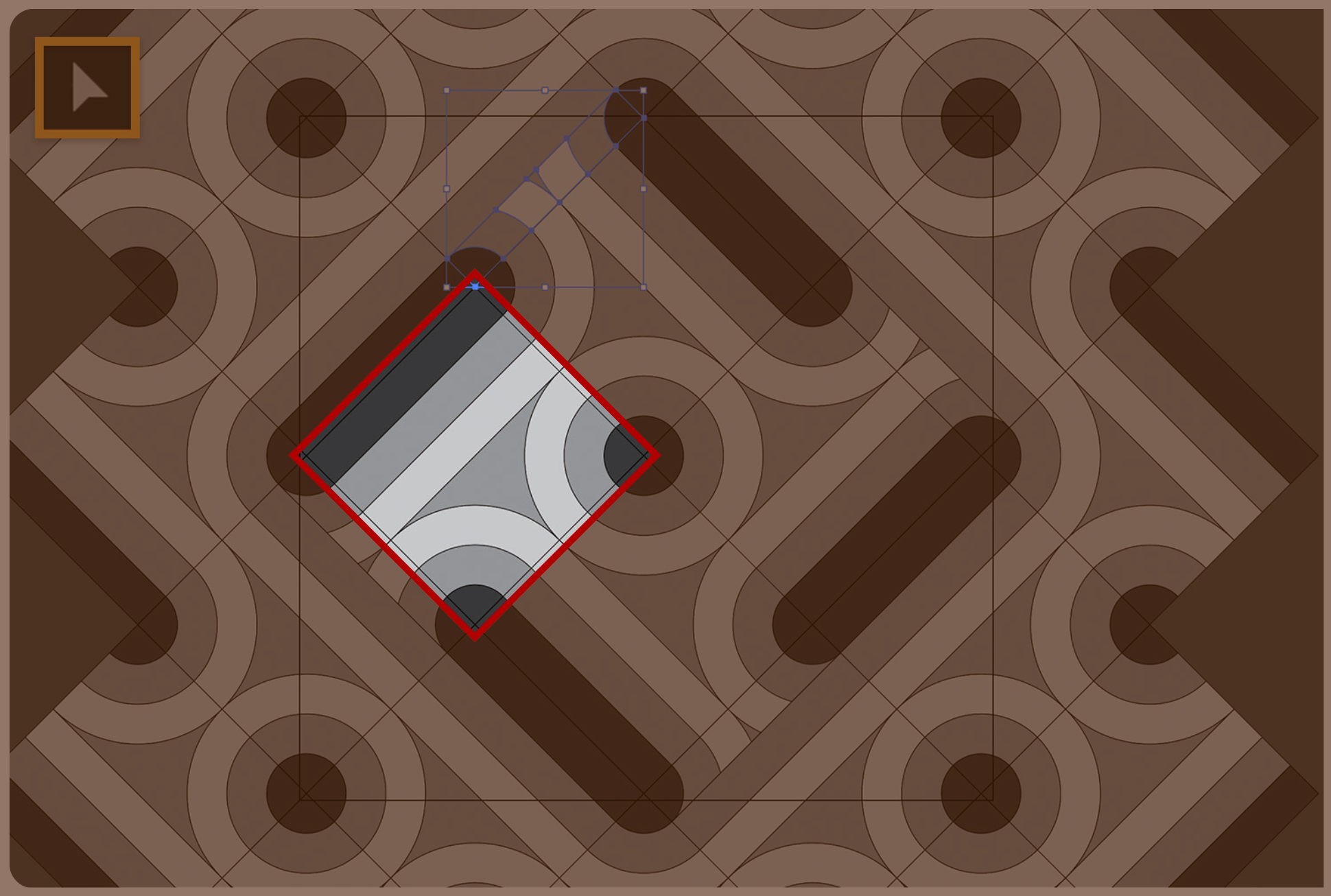}}\par\smallskip\scriptsize\textbf{Reference construction.}\ ``Create the base unit tile for the gray-toned De\,Nigris `3/4 Decorative' geometric tiling shown above.'' &
  \adjustbox{valign=t}{\includegraphics[width=\linewidth]{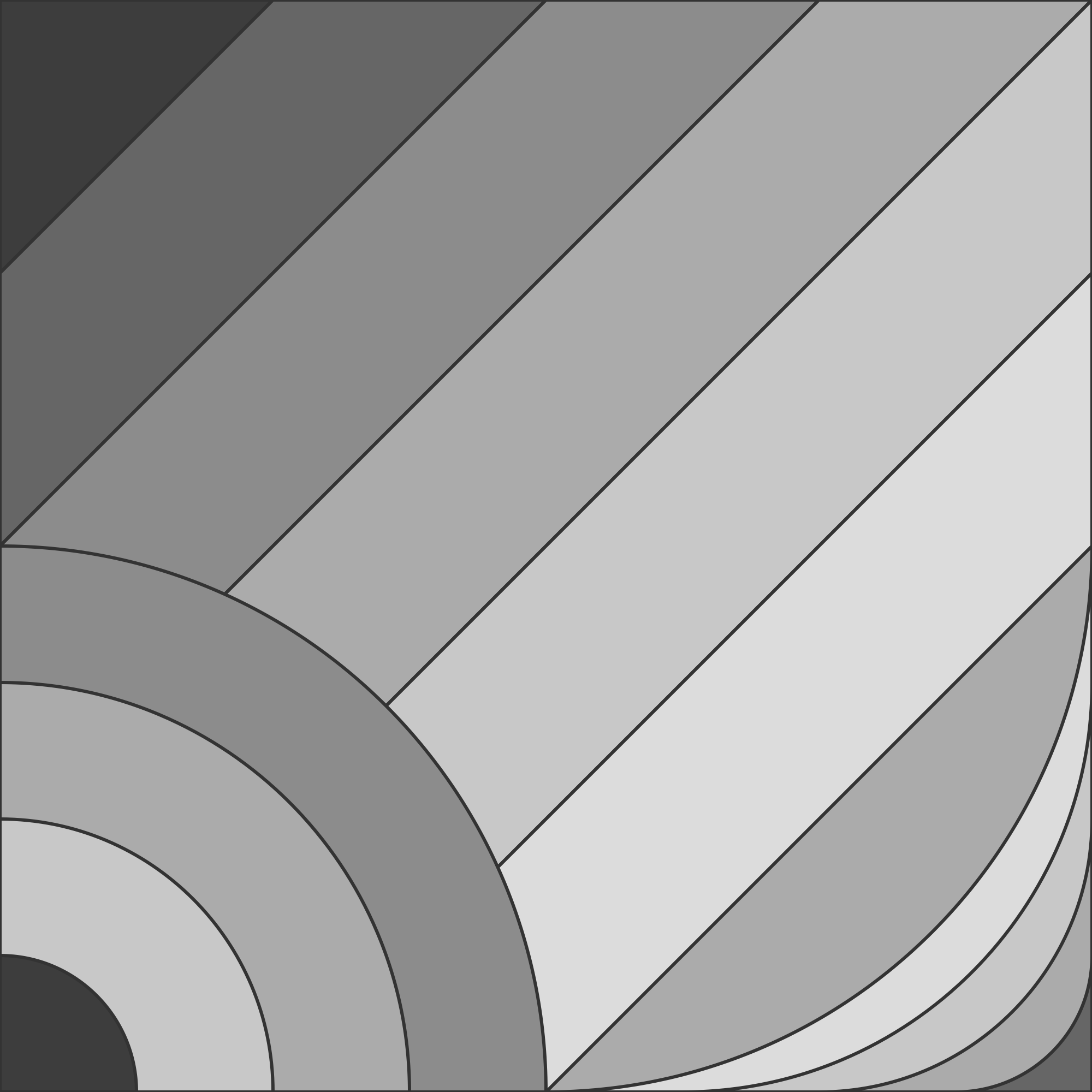}} &
  \adjustbox{valign=t}{\includegraphics[width=\linewidth]{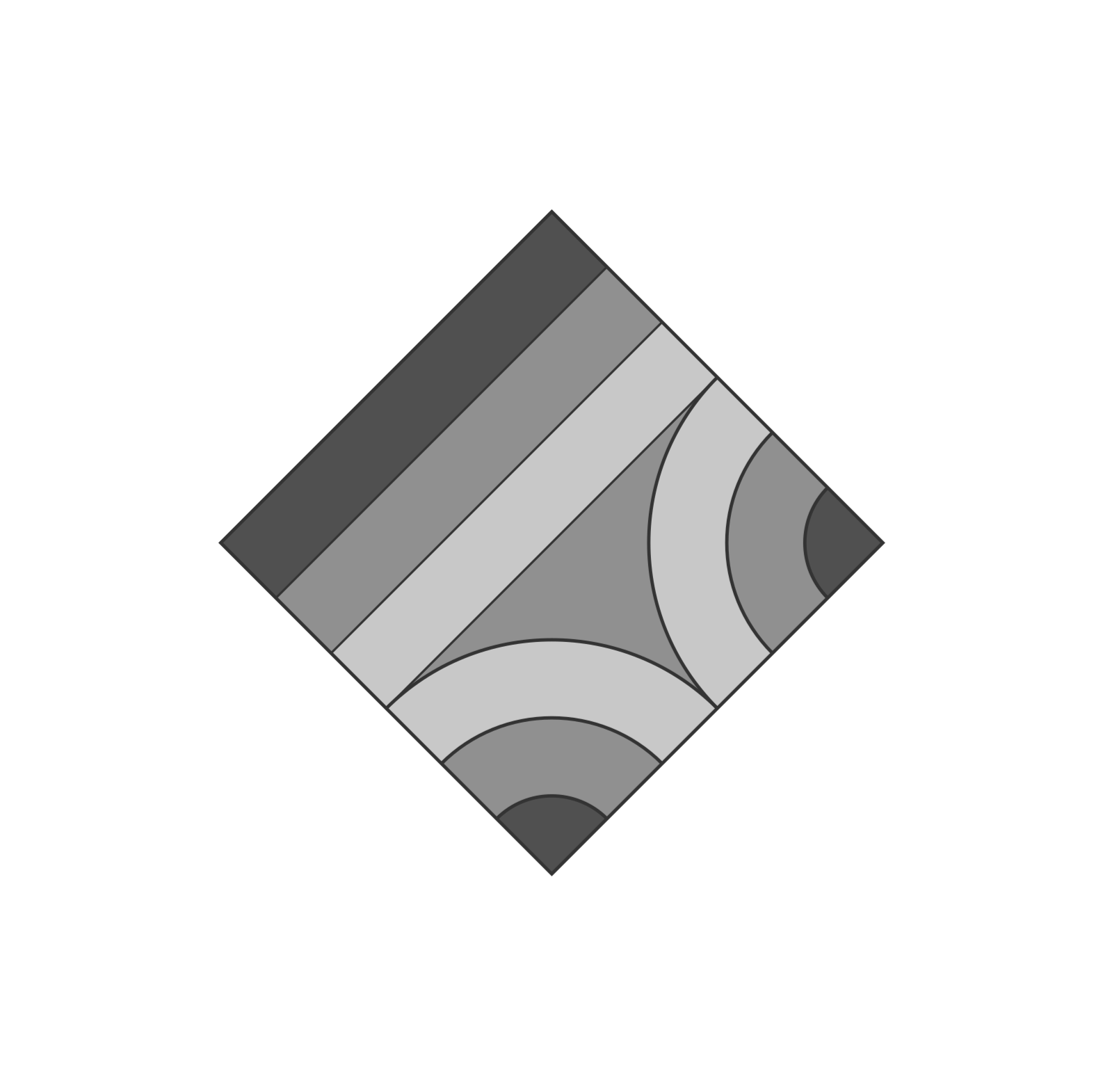}} \\
\end{tabularx}
\caption{\textbf{Personalizing procedures to a user's method and target.} Each case compares the same frozen backbone without the personalized skill (left) and with it (right). Top: preference-aware retrieval and tool scoping reproduce the user's layered snow technique. Middle: skill guidance improves the nested-frame construction, while remaining geometric misalignment motivates numeric verification and explicit placement operations. Bottom: a skill distilled from the user's construction produces an editable base tile for the illustrated reference pattern. These user-guided cases are separate from the shared-bank experiments.}
\label{fig:personalization}
\end{figure}

Whereas the shared bank captures general-purpose workflows recurring across traffic, creative professionals often insist on a specific \emph{construction method} or exact geometric structure. Here we examine three user-guided case studies (Figure~\ref{fig:personalization}), each comparing the same frozen backbone with and without a personalized \code{SKILL.md} distilled from interactive user guidance. All personalized skills are evaluated in isolation and excluded from the shared bank and main experiments.

\paragraph{Method preference.}
To add falling snow to a night-street photo, the user specifies a classic multi-layer photographic recipe (fill, add noise, Gaussian blur, threshold, and screen-blend, repeated across depth planes), whereas the agent's unguided default invokes a single particle-field tool. Reproducing the user's technique requires preference-aware retrieval and tool scoping (Appendix~\ref{sec:bank}) to disable the particle shortcut, yielding the layered atmospheric depth in Figure~\ref{fig:personalization} (top).

\paragraph{Geometric precision and numeric verification.}
For a three-layer Droste effect inside a photographed picture frame, skill guidance substantially improves recursive frame placement over \textsc{Base}, though slight boundary drift remains visible (Figure~\ref{fig:personalization}, middle). Because VLM judges often miss sub-pixel or narrow-band seam misalignments, deterministic pixel-band checks around frame borders and explicit coordinate placement complement natural-language skill instructions on precision-critical layouts.

\paragraph{Reference geometric construction.}
Asked to build the base unit tile of a De\,Nigris ``3/4 Decorative'' geometric mosaic, \textsc{Base} produces an inaccurate diagonal approximation, whereas a skill distilled from user-guided construction reproduces the intended editable vector tile with intact paths, arcs, and fills ready for tessellation (Figure~\ref{fig:personalization}, bottom).

\begin{figure}[t]
\centering
\setlength{\tabcolsep}{0pt}
\begin{tabularx}{\linewidth}{@{}>{\centering\arraybackslash}X@{\hspace{8pt}}>{\centering\arraybackslash}X@{}}
\adjustbox{valign=t}{\includegraphics[width=\linewidth]{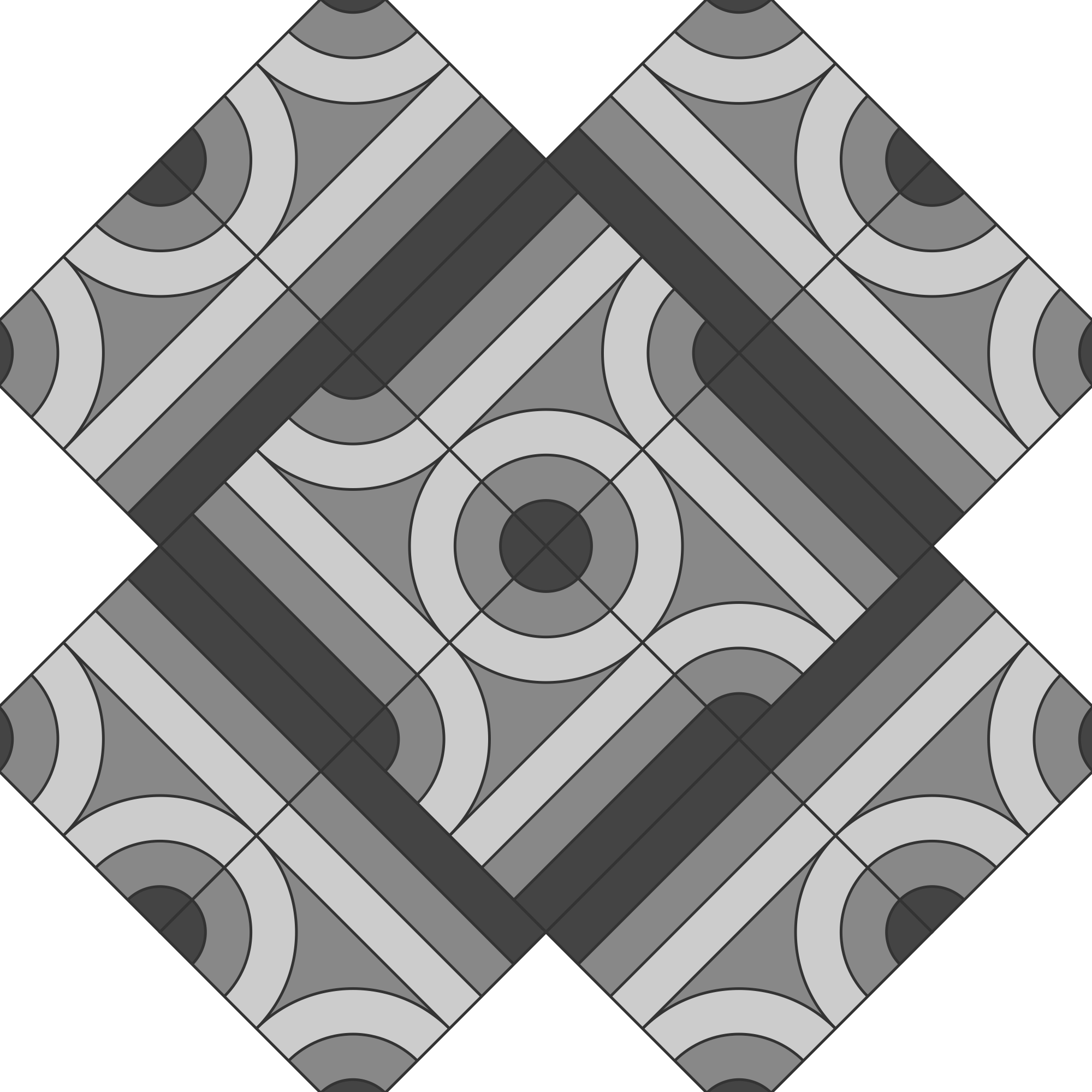}} &
\adjustbox{valign=t}{\includegraphics[width=\linewidth]{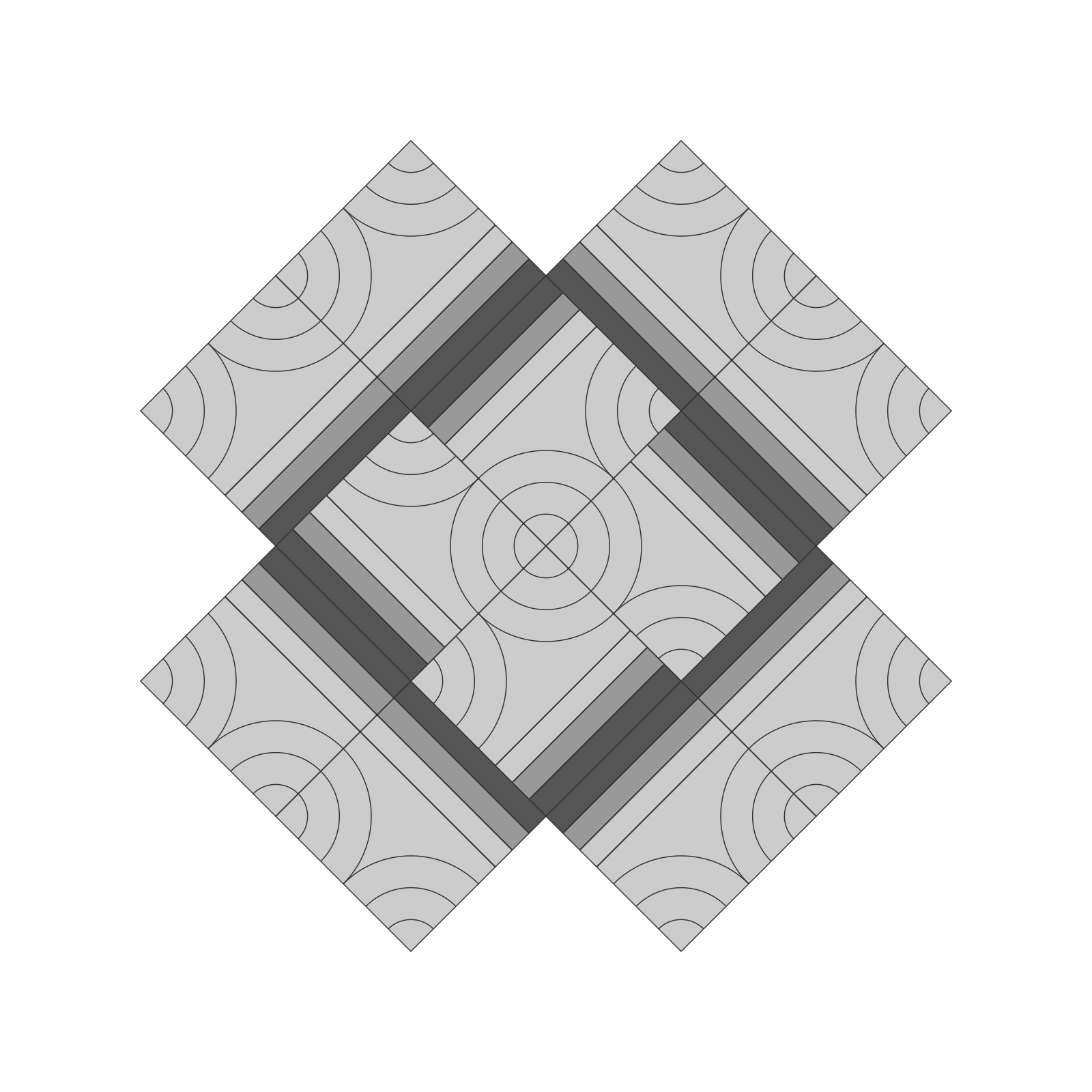}} \\
\small (a) Agent, under human guidance & \small (b) Agent, using the annotated skill \\
\end{tabularx}
\caption{\textbf{Extending a unit procedure to a full construction.} (a) The full reference pattern produced under human guidance. (b) Execution of the distilled skill preserves the overall tiling layout while omitting the three-tone fills and fine structure. The comparison identifies a role for explicit orchestration when composing a short, reusable procedure into a longer design.}
\label{fig:full-pattern}
\end{figure}

\paragraph{Scaling unit skills to full-canvas orchestrations.}
Figure~\ref{fig:full-pattern} extends the single mosaic tile to a full multi-tile tessellation whose human-guided demonstration spans hundreds of turns. Executing a single distilled skill preserves the global grid arrangement but drops three-tone shading and fine inner arcs. Short unit skills and explicit macro-level orchestration thus play complementary roles: a \code{SKILL.md} captures the reusable unit construction, while deterministic layout loops or hierarchical plans should govern large-scale repetition.

\section{Discussion: Product Adaptation and Scope}
\label{app:product-discussion}

\paragraph{Where adaptation lives when weights are frozen.}
In our framework, all continual adaptation resides in human-readable \code{SKILL.md} files, covering both their trigger \code{description} fields (which govern lexical routing) and their step-by-step tool instructions (which govern execution). Keeping foundation models, tools, and the replay harness fixed offers two practical advantages for creative software products: updates remain transparent to product engineers, and problematic skills can be version-rolled back or edited individually without retraining.

\paragraph{Why local admission needs whole-bank monitoring.}
Our matched replay gate verifies local non-regression on sampled briefs from a skill's own cluster or retrieval history, keeping admission compute tractable. However, adding or rewriting a skill's \code{description} can subtly shift lexical retrieval rankings on broader traffic, as illustrated by Round 4's slight dip at low completeness thresholds before Round 5's deepening pass consolidated those skills (Figure~\ref{fig:retention}). Pairing inexpensive local replay gates with periodic whole-bank held-out tracking provides a practical two-level safeguard.

\paragraph{From shared workflows to user-specific methods.}
While the shared bank captures general-purpose workflows recurring across users, individual designers often care about \emph{how} an effect is constructed (e.g., layered noise/blur snow vs.\ particle shortcuts) or require exact geometric nesting beyond VLM perceptual resolution. Appendix~\ref{app:personalization} shows that the same \code{SKILL.md} abstraction extends to user-specific workflows when paired with preference-conditioned retrieval, tool scoping, numeric geometric checks, and macro-level orchestration for hundred-turn constructions.

\section{Gallery on the Held-Out Prompt Set}
\label{app:gallery}

\begin{figure}[p]
  \centering
  \includegraphics[width=\linewidth,height=0.92\textheight,keepaspectratio]{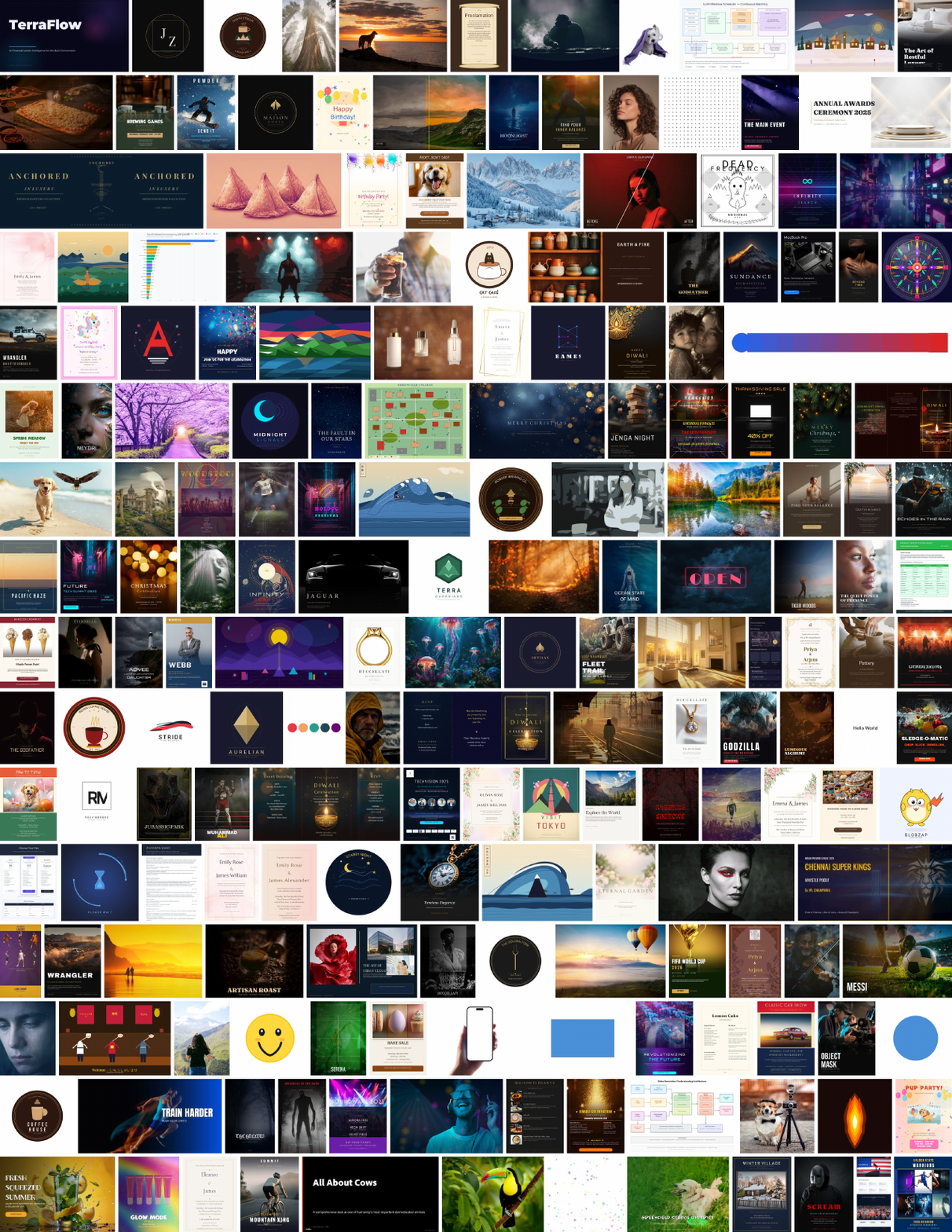}
  \caption{%
    \textbf{Opus~4.6 backbone (low thinking).} The gallery shows all
    $193$ held-out prompts rendered with the round-5 skill bank, in random order. Prompt text is withheld
    (internal test set), while aspect ratios are preserved and no panel is selected for
    quality.%
  }
  \label{fig:gallery-opus}
\end{figure}

\begin{figure}[p]
  \centering
  \includegraphics[width=\linewidth,height=0.92\textheight,keepaspectratio]{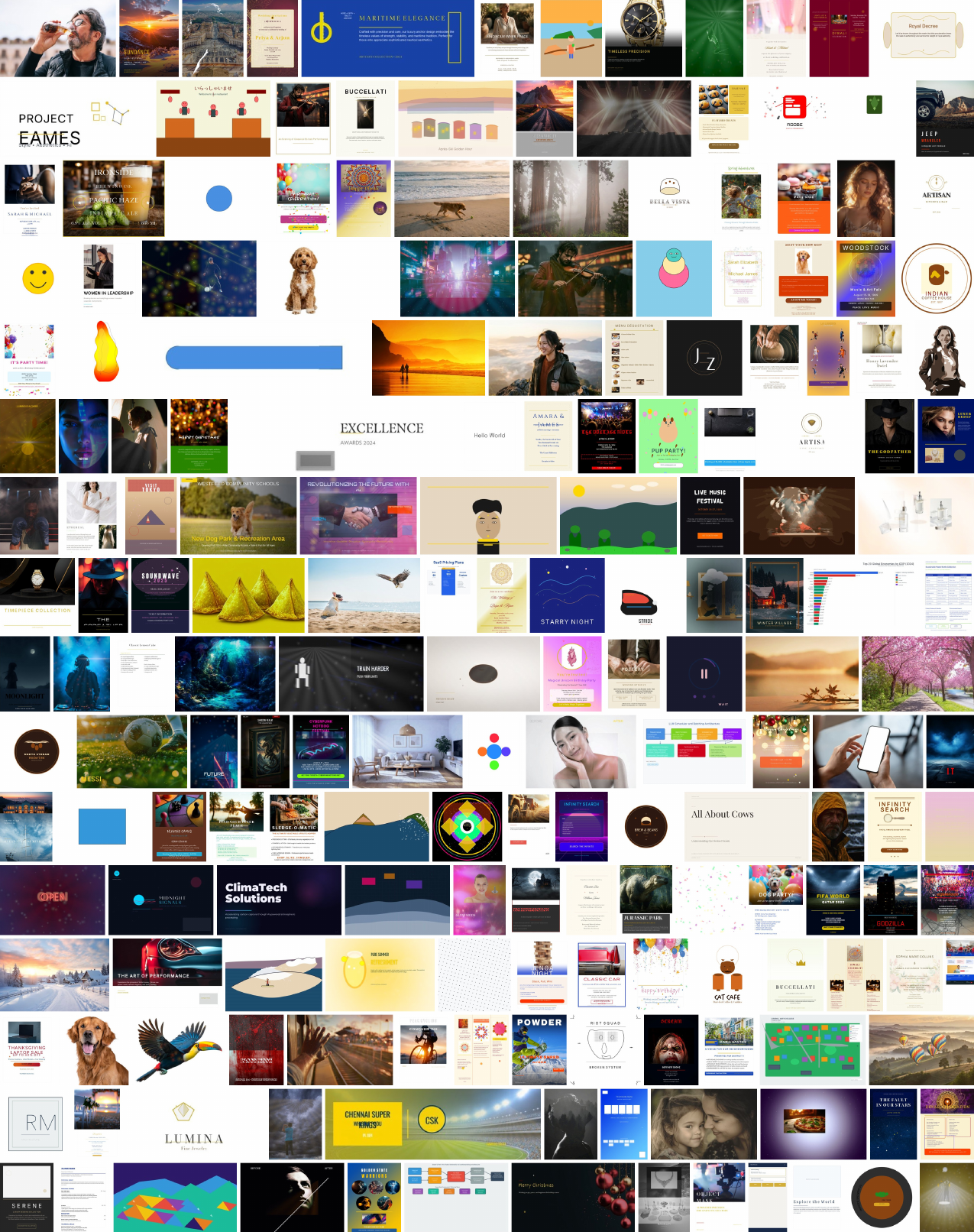}
  \caption{%
    \textbf{Sonnet~4 backbone (low thinking).} The same $193$ held-out prompts as
    Figure~\ref{fig:gallery-opus}, rendered under the identical protocol.%
  }
  \label{fig:gallery-sonnet}
\end{figure}

Figures~\ref{fig:gallery-opus} and~\ref{fig:gallery-sonnet} show the designs our
system produces on the held-out prompt set, one panel per prompt, for the Opus~4.6
and Sonnet~4 backbones respectively, both run with a low thinking budget. The two
figures cover the same $193$ prompts rendered with the round-5 skill bank.
The corresponding no-skill baselines and side-by-side pairs are omitted here
for space. The prompts span the full range of the benchmark (posters, flyers,
invitations, logos, brand and UI layouts, photographic composites, and low-level
primitive requests) and vary widely in canvas aspect ratio, which is preserved for
every panel: images are uniformly scaled and tiled into rows of equal height, never
cropped or stretched.

The benchmark is an internal test set, and our content policy does not
permit releasing the prompt text.
Within that split nothing further is selected: both galleries show every prompt, in
a fixed random order, so readers can assess typical output quality instead of a hand-picked subset. Individual panels are
necessarily small at this density and are meant to convey the aggregate
distribution of quality, layout structure, and stylistic variety.

% --- Opus 4.6 gallery -------------------------------------------------------
% Fit the complete mosaic on one page, leaving room for the caption.

% --- Sonnet 4 gallery -------------------------------------------------------

\section{Ethical Statement}
\label{app}

This work studies an offline evolution and replay pipeline for a graphic-design agent. The evolution data include design briefs derived from pre-existing user traffic collected in the course of normal product use, together with LLM-augmented variants. No users have been recruited and no additional user data or annotations have been collected specifically for this study to date. We plan to conduct a small-scale user study to evaluate the effectiveness of \textsc{Evolve} from a human perspective. The held-out internal benchmark is human-authored, and the main experiments use automated evaluation rather than human reward labels. User-guided personalization examples are evaluated separately and are excluded from the shared skill pool and all main-paper experiments.

The use of pre-existing user data raises privacy and confidentiality considerations. User-derived data were accessed and processed within the organization's established data-governance and access-control framework. User-derived inputs were filtered and processed to reduce the exposure of personally identifiable or other sensitive information. We do not release user-derived prompts or trajectories. Because the evaluation benchmark is internal, we also withhold its full prompt set and trajectory contents. The qualitative gallery contains only rendered outputs for 193 held-out prompts, without the corresponding prompt text.

The system uses hosted proprietary language-model services through Amazon Bedrock and Azure, locally served models, and assets retrieved from Adobe Stock. User-derived data may be processed by these hosted model services under the organization's enterprise agreements with the respective providers, which prohibit the use of such data for training the providers' models. Third-party assets and user-derived materials are not publicly released except where such disclosure is authorized and consistent with applicable licenses and data-use requirements.

Finally, graphic-design automation can be misused to create misleading, deceptive, or rights-infringing content. Deploying such systems therefore requires content safeguards, access controls, and human oversight.

\end{document}